\documentclass[10pt,twocolumn,letterpaper]{article}

\usepackage[pagenumbers]{wacv}

\usepackage{array}
\usepackage{amsfonts}
\usepackage{float}
\usepackage{multirow}

\graphicspath{{images/}}

\definecolor{wacvblue}{rgb}{0.21,0.49,0.74}
\usepackage[pagebackref,breaklinks,colorlinks,allcolors=wacvblue]{hyperref}

\def\confName{WACV}
\def\confYear{2027}

\title{XDG: Accelerated Visual Disambiguation}

\author{
Gonglin Chen\textsuperscript{1,2} \quad
Ben Southall\textsuperscript{3} \quad
Hanyuan Xiao\textsuperscript{1,2} \quad
Wenbin Teng\textsuperscript{1,2}\\
Haolin Xiong\textsuperscript{1,2} \quad
Tianwen Fu\textsuperscript{1,2} \quad
Junyi Ouyang\textsuperscript{1,2} \quad
Kshitij Singh Minhas\textsuperscript{3}\\
Supun Samarasekera\textsuperscript{3} \quad
Rakesh Kumar\textsuperscript{3} \quad
Yajie Zhao\textsuperscript{1,2}\\[3pt]
\textsuperscript{1}USC Institute for Creative Technologies \quad
\textsuperscript{2}University of Southern California \quad
\textsuperscript{3}SRI International
}

\begin{document}

\makeatletter
\twocolumn[{
    \@maketitle
    \centering
    \includegraphics[width=0.98\textwidth]{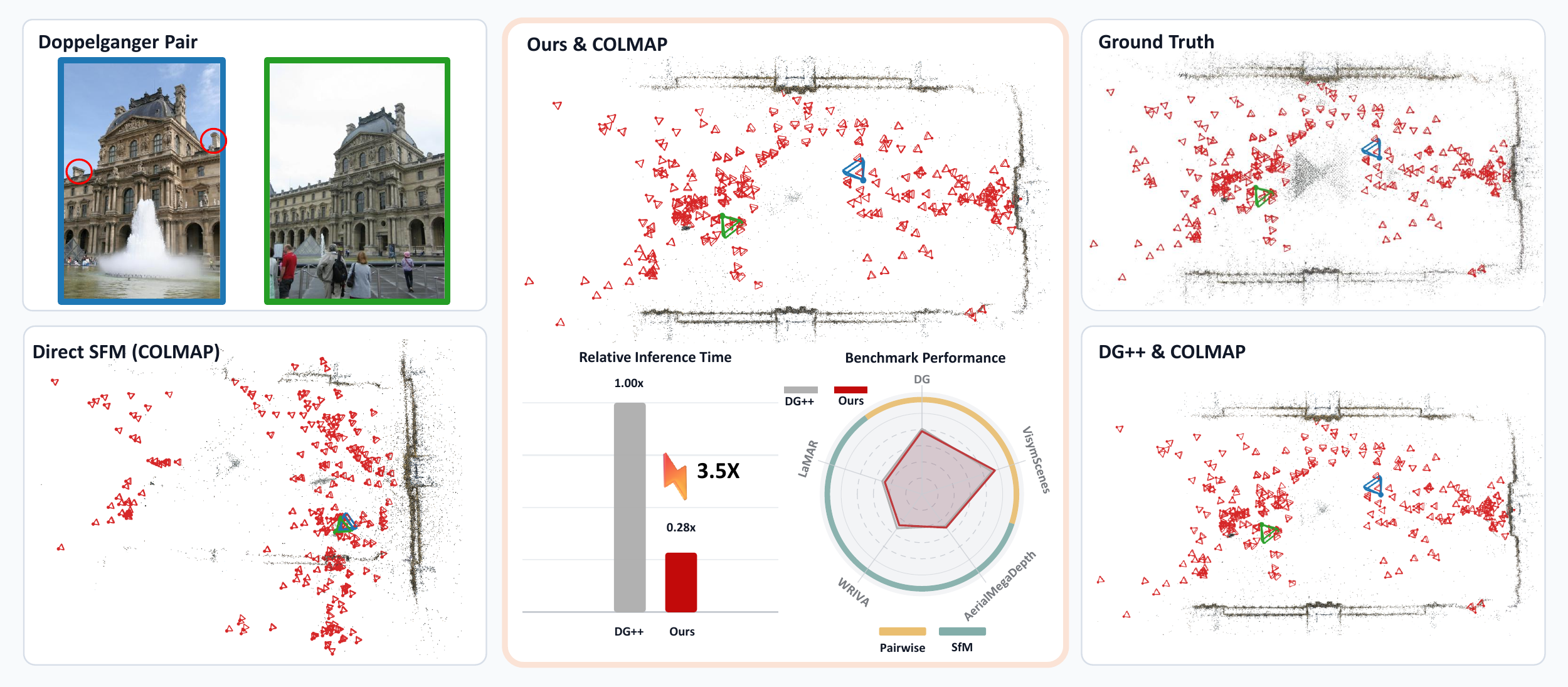}
    \captionof{figure}{\textbf{XDG efficiently removes visually plausible false matches while preserving reconstruction quality.} Top left: an example of a doppelganger pair from two distinct parts of a scene (the louvre museum) from AerialMegaDepth~\cite{vuong2025aerialmegadepth}. The images share similar visual structure, but the circled details reveal inconsistent local geometry and appearance. The highlighted blue and green camera poses show where this pair is placed in each reconstruction. A vanilla COLMAP reconstruction is corrupted by multiple similar false matches, while XDG filters doppelganger edges and recovers a camera layout comparable to Doppelgangers++ (DG++)~\cite{xiangli2025doppelgangers} and close to ground truth. Across pairwise and SfM benchmarks, XDG maintains comparable disambiguation performance compared to DG++ while significantly reduce the inference cost.}
    \label{fig:teaser}
    \vspace{1em}
}]
\makeatother

\begin{abstract}
Visual aliasing, also known as the doppelganger problem, remains a key challenge for structure-from-motion (SfM): visually similar but physically distinct surfaces can produce incorrect image matches and degrade reconstruction quality. Previous work mitigates this issue with geometry-aware foundation-model features, but places a heavy transformer classifier on top of the backbone, making large-scale disambiguation expensive. We introduce XDG, an efficient visual disambiguation model designed for scalable SfM. Our key observation is that a 3D foundation model already performs the cross-view geometric reasoning necessary for visual disambiguation, so doppelganger classification should adapt the backbone representation directly rather than relearn pair reasoning in a separate heavy decoder. XDG fine-tunes Depth Anything 3 with lightweight LoRA adapters and repurposes its camera tokens as compact pair-level classification tokens. A compact MLP head predicts whether a candidate image pair observes the same 3D surface. Extensive experiments show that XDG provides a favorable accuracy--efficiency tradeoff: it remains competitive with the state-of-the-art disambiguation method across pairwise and reconstruction benchmarks and delivers more than a $3\times$ inference speedup. On individual LaMAR scenes containing thousands of images, XDG saves more than $10$ hours of visual disambiguation processing. Code is available at \href{https://github.com/xtcpete/xdg}{https://github.com/xtcpete/xdg}.

\end{abstract}

%-------------------------------------------------------------------------
\vspace{-1 em}
\section{Introduction}
\label{sec:intro}
Accurate 3D reconstruction from unordered image collections is fundamental to computer vision and increasingly critical for producing the geometric supervision required by emerging 3D foundation models. Modern structure-from-motion (SfM) systems rely on robust image retrieval, local feature matching, and geometric verification to build match graphs before reconstruction~\cite{schonberger2016structure}. Despite these advances, SfM pipelines remain vulnerable to visual aliasing: distinct physical surfaces can share highly similar appearance, causing false image matches that pass local matching and contaminate the reconstruction. This failure mode is formalized as the doppelganger problem~\cite{cai2023doppelgangers}. If such edges remain in the match graph, SfM can merge unrelated structures, register cameras incorrectly, or produce fragmented and distorted reconstructions, as shown in Fig.~\ref{fig:teaser}.
A practical disambiguation system must therefore satisfy two competing requirements. It must be geometrically informed enough to distinguish true overlap from repeated appearance, while also being efficient enough to run on large SfM graphs containing thousands of candidate image pairs. Early work addressed this problem with a CNN-based binary classifier trained on labeled doppelganger pairs, where tentative correspondences from LoFTR~\cite{sun2021loftr} were used to estimate an affine warp before classification~\cite{cai2023doppelgangers}. Doppelgangers++ (DG++)~\cite{xiangli2025doppelgangers} improved generalization by leveraging MASt3R~\cite{leroy2024grounding}, a large 3D model, and then training transformer-based classifiers~\cite{vaswani2017attention} on its features. DG++ already disambiguates many challenging cases effectively; its runtime, rather than its accuracy, becomes the practical bottleneck at scale. Its geometry-aware backbone and heavy transformer classifiers must be evaluated over every candidate edge, so processing scenes with thousands of images can require tens of hours.

In this work, we introduce XDG, an efficient geometry-aware doppelganger classifier designed for scalable SfM disambiguation. Our key idea is to adapt the 3D backbone directly and classify from its native representation with a lightweight classifier. XDG uses Depth Anything 3 (DA3)~\cite{lin2025depth} as the geometry-aware backbone and inserts low-rank adaptation (LoRA) modules~\cite{hu2022lora} into its attention and feed-forward projections. Instead of training a separate transformer decoder, XDG repurposes DA3's stage-wise camera tokens as compact pair-level classification tokens. These tokens already participate in DA3's alternating local and global feature interactions, so after LoRA fine-tuning, they efficiently summarize the relationship between views. A small MLP head then predicts whether the pair is a true match or a doppelganger edge.

This design differs from prior geometry-aware disambiguation in three ways. First, XDG adapts the foundation model itself to the doppelganger disambiguation task using parameter-efficient LoRA updates, while keeping the pretrained DA3 weights frozen. Second, XDG removes the heavy post-backbone transformer classifier and performs classification directly from compact camera tokens. Third, XDG evaluates both image orders for robustness to input order, but fuses the resulting tokens before classification, avoiding the two separate order-specific classification heads used in prior work~\cite{xiangli2025doppelgangers}. Importantly, our ablation in Tab.~\ref{tab:xdg_ablation} shows that backbone replacement alone is insufficient: using DA3 with a DG++-style dense-token transformer classifier provides only limited acceleration. The main efficiency gain instead comes from the proposed modules, which classify image pairs directly from the backbone’s native camera tokens, enforce order consistency through symmetric token aggregation, and use a lightweight MLP head for final prediction. 

We evaluate XDG on both pairwise visual disambiguation and downstream reconstruction. On the DG~\cite{cai2023doppelgangers} and VisymScenes~\cite{xiangli2025doppelgangers} pairwise benchmarks, XDG achieves accuracy comparable to the state-of-the-art method DG++~\cite{xiangli2025doppelgangers} while reducing per-pair inference time by more than $3\times$. Across multiple reconstruction benchmarks, including the unseen WRIVA~\cite{cjk5-gf33-24} and LaMAR~\cite{sarlin2022lamar} datasets, XDG is competitive overall and significantly reduces total processing time. On the LaMAR scenes with $7.5$K--$9.3$K images, XDG saves $10.85$--$13.91$ hours of visual-disambiguation processing per scene compared with DG++. In summary, our contributions are:
\begin{enumerate}[leftmargin=*, itemsep=0pt, topsep=0pt]
\item We propose XDG, an efficient geometry-aware classifier for removing doppelganger edges from SfM match graphs.
\item We show that doppelganger classification does not require a heavy transformer classifier on top of a frozen 3D foundation model; instead, parameter-efficient LoRA adaptation with a simple MLP head can directly adapt the backbone to the task.
\item We demonstrate that XDG achieves competitive pairwise disambiguation and downstream SfM reconstruction performance while substantially reducing disambiguation cost, a key advantage for large-scale scene reconstruction.
\end{enumerate}

\begin{figure*}[t]
    \centering
    \includegraphics[width=0.96\linewidth]{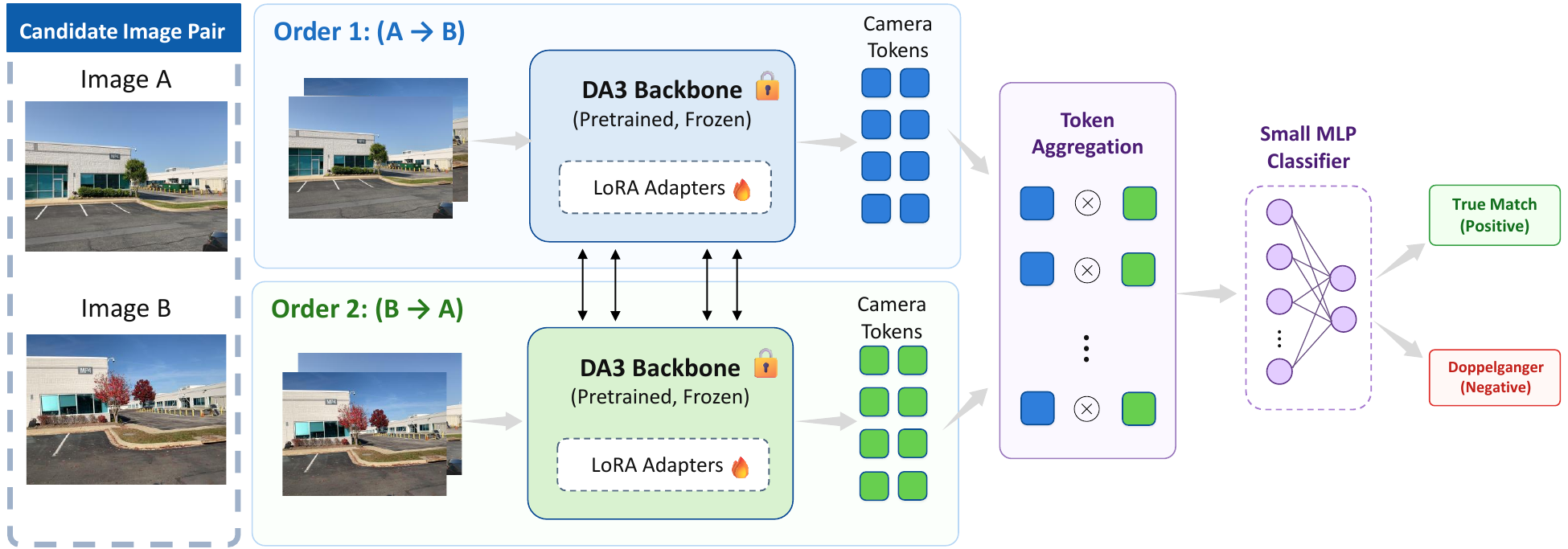}
    \caption{\textbf{XDG architecture.} Given a candidate pair $(I_A,I_B)$, XDG processes both image orders using a shared DA3-Base backbone with LoRA adapters. Camera tokens from the reversed pass are realigned with the canonical image order and fused stage-wise. The fused tokens are projected, normalized, averaged across stages and views, and then classified by a lightweight MLP as either a true match or a doppelganger edge.}
    \label{fig:xdg_architecture}
\vspace{-1 em}
\end{figure*}

\section{Related Work}
\label{sec:related}
\textbf{Local Feature Matching.} Classical structure-from-motion (SfM) pipelines such as COLMAP~\cite{schonberger2016structure} rely on local feature matching to establish image correspondences. Traditionally, hand-crafted descriptors such as SIFT~\cite{lowe2004distinctive} have been widely used for matching keypoints across image pairs. More recently, learning-based approaches have substantially improved matching robustness and coverage, leading to stronger reconstruction performance~\cite{chen2025geometry,chen2025rdd,sun2021loftr,he2024dfsfm,lindenberger2023lightglue,sarlin2020superglue,detone2018superpoint,yi2016lift,rocco2018neighbourhood,leroy2024grounding}. However, these methods are primarily optimized to maximize the number of repeatable correspondences between visually similar regions, often using supervision derived from image overlap or geometric consistency. As a result, their training data may contain visually aliased or doppelganger patterns, while their objectives provide limited incentive to reject such false matches. Moreover, local matchers typically operate at the patch or keypoint level, without explicitly modeling global image context or higher-level 3D scene consistency. Consequently, although highly effective for standard correspondence estimation, they remain vulnerable to producing spurious matches between distinct yet visually similar surfaces, particularly in doppelganger image pairs.

\textbf{3D Learning.}
Recent advances in learning-based 3D reconstruction have produced a wave of methods that replace traditional keypoint matching and iterative optimization with feed-forward neural networks. DUSt3R~\cite{wang2024dust3r} introduced this paradigm by regressing dense point maps from pairs of uncalibrated images, enabling direct recovery of scene geometry and camera poses. Building on this direction, recent methods train large feed-forward transformers to reconstruct 3D geometry from one or many views in a single forward pass: VGGT~\cite{wang2025vggt} jointly predicts camera parameters, depth maps, point maps, and feature tracks; $\pi^3$~\cite{wang2025pi} improves robustness with a permutation-equivariant architecture that removes dependence on a fixed reference view; and Depth Anything 3~\cite{lin2025depth} adopts a unified design for consistent any-view geometry reconstruction. Despite their strong performance, purely feed-forward methods still struggle to match the precision and global consistency required by large-scale SfM pipelines, particularly when image collections contain repeated or visually aliased structures~\cite{pan2026gluemap}. Their computational and memory costs also grow rapidly with the number of views, making them difficult to use directly as scalable reconstruction systems. More recently, GLUEMAP~\cite{pan2026gluemap} addresses these limitations by combining feed-forward local reconstruction with classical global motion averaging and bundle adjustment, while using DG++~\cite{xiangli2025doppelgangers} as a two-view disambiguation stage to remove harmful doppelganger edges. XDG complements this hybrid reconstruction framework by replacing the costly DG++ stage with a faster classifier that preserves competitive reconstruction accuracy in most settings.

\textbf{Disambiguation in SfM and Image Matching.}
Visual disambiguation addresses a failure mode that is complementary to feature matching: instead of only finding correspondences between images, it must decide whether visually plausible correspondences actually arise from the same 3D structure. Classical SfM systems reduce false matches through geometric verification and robust optimization~\cite{schonberger2016structure}, but repeated structures and near-duplicate appearances can still survive these checks and produce incorrect image edges. Earlier ambiguity-handling methods reasoned over the structure of the image graph, loop constraints, duplicate scene components, missing correspondences, or geodesic context~\cite{zach2008missing,zach2010disambiguating,jiang2012seeing,wilson2013network,heinly2014correcting,yan2017distinguishing}; however, these approaches rely largely on hand-designed graph or correspondence cues rather than directly learning from image content. Doppelgangers~\cite{cai2023doppelgangers} first formulated this problem explicitly as doppelganger detection and trained a CNN-based binary classifier to reject ambiguous image pairs before reconstruction, improving SfM robustness but remaining limited by an appearance-centric representation. Doppelgangers++~\cite{xiangli2025doppelgangers} improves visual disambiguation by expanding the training data and using 3D-aware MASt3R~\cite{leroy2024grounding} features with transformer-based classifiers, achieving stronger performance and integrating with both standard SfM and MASt3R-SfM~\cite{duisterhof2025mast3r} pipelines. However, its reliance on a heavy 3D foundation model plus duplicated transformer classifiers increases computational cost, especially when large image collections require evaluating many candidate edges. Our work follows the same direction but takes a different efficiency route: we fine-tune DA3 with LoRA and classify directly from its camera tokens, avoiding an expensive transformer reasoning module after the backbone.

\vspace{-1 em}

\section{Method}

\label{sec:method}

Given a candidate image pair $(I_A, I_B)$ from an SfM match graph, our goal is to decide whether the pair observes a common 3D surface or forms a doppelganger edge that should be pruned before reconstruction. Following~\cite{cai2023doppelgangers,xiangli2025doppelgangers}, we formulate this as binary classification, where true matches are positives and ambiguous or non-overlapping pairs are negatives. XDG accelerates this task by removing the heavy post-backbone transformer classifier.

\subsection{Model Architecture}
\label{sec:model_architecture}

As shown in Fig.~\ref{fig:xdg_architecture}, XDG consists of three main components: a DA3 backbone with LoRA adapters~\cite{hu2022lora}, a symmetric token aggregation module, and a lightweight classification head. Given a candidate image pair $(I_A, I_B)$, we apply the same LoRA-adapted DA3 encoder to both image orders, $(I_A, I_B)$ and $(I_B, I_A)$. From each pass, XDG extracts stage-wise camera tokens, realigns tokens corresponding to the same image, and fuses them into an order-consistent pair representation. The fused tokens are then projected, average-pooled, and passed to a small MLP to predict whether the pair is a true match or a doppelganger.

\textbf{DA3 backbone with LoRA adaptation.}
XDG uses the Base variant of Depth Anything 3 (DA3)~\cite{lin2025depth} as its geometry-aware backbone. DA3 alternates local and global attention over multi-view image tokens, providing the cross-view reasoning needed for doppelganger detection. We freeze the pretrained DA3 weights and insert low-rank adaptation modules~\cite{hu2022lora} into selected linear layers of the backbone.

For a pretrained linear projection $W$, LoRA adds a trainable low-rank residual:
\begin{equation}
    y = Wx + \frac{\alpha}{r} B A x,
\end{equation}
where $A \in \mathbb{R}^{r \times d_{\mathrm{in}}}$ and $B \in \mathbb{R}^{d_{\mathrm{out}} \times r}$ are trainable, $r$ is the LoRA rank, and $\alpha$ is a scaling factor. The original projection $W$ remains frozen. In our implementation, LoRA is applied to the DA3 attention and feed-forward projections \texttt{qkv}, \texttt{proj}, \texttt{fc1}, and \texttt{fc2}. We use rank $r=8$, $\alpha=16$, and dropout $0.05$. The DA3 camera-token parameters are also left trainable, while the rest of the pretrained backbone parameters remain frozen.

\textbf{Repurposing the camera tokens.}
DA3 maintains camera tokens that are injected into the transformer stream at the start of the alternating cross-view blocks for each input image. These tokens are designed to participate in global multi-view reasoning and therefore provide a natural compact representation across images. For doppelganger detection, we repurpose them as classification tokens.

Let $\mathcal{S}=\{5,7,9,11\}$ denote the four DA3 stages used by XDG, and let $C_e=1536$ be the camera-token dimension. For the ordered pair $(I_A,I_B)$, the LoRA-adapted DA3 backbone returns
\begin{equation}
    \{(F_s^{A \rightarrow B}, C_s^{A \rightarrow B})\}_{s \in \mathcal{S}}
    =
    \Phi_{\theta}(I_A,I_B),
\end{equation}
where $F_s^{A \rightarrow B}$ denotes the dense patch tokens and $C_s^{A \rightarrow B}\in\mathbb{R}^{2\times C_e}$ contains one camera token for each image. The same encoder $\Phi_\theta$ also processes the reversed pair $(I_B,I_A)$ to produce $C_s^{B \rightarrow A}$ by batching the two image orders together. Unlike DG++~\cite{xiangli2025doppelgangers}, XDG does not pass the dense tokens to a heavy transformer classifier; it classifies the pair using only the compact camera tokens.

\textbf{Symmetric token aggregation.}
Because the ordering of an image pair is arbitrary, XDG combines the camera tokens from both input orders. Let $\pi$ swap the two-view dimension of the reversed output, thereby restoring the canonical $(A,B)$ view order. At each stage, XDG computes
\begin{equation}
    \widetilde{C}_s =
    g_C\left(
    \left[C_s^{A\rightarrow B};
    \pi(C_s^{B\rightarrow A})\right]\right),
\end{equation}
where $[\cdot\,;\cdot]$ denotes concatenation along the feature dimension and $\widetilde{C}_s\in\mathbb{R}^{2\times C_e}$. The same two-layer fusion MLP $g_C$ is shared across all stages and maps $2C_e\rightarrow4C_e\rightarrow C_e$, with a GELU activation between its linear layers.

\textbf{Classification head.}
The classification head independently projects the fused tokens from each stage to a hidden dimension $C_d$ and applies layer normalization:
\begin{equation}
    T_s = \operatorname{LN}_s(W_s^C \widetilde{C}_s).
\end{equation}
It then averages the projected tokens over the four stages and two views:
\begin{equation}
    t = \frac{1}{2|\mathcal{S}|}
    \sum_{s\in\mathcal{S}}\sum_{v\in\{A,B\}}T_{s,v}.
\end{equation}
Finally, it predicts two logits:
\begin{equation}
    z = h(t),
\end{equation}
where $h$ is a layer-normalized MLP mapping $C_d\rightarrow C_h\rightarrow2$, with GELU activation and dropout between its linear layers. The two logits correspond to the doppelganger (false-match) class and the true-match class, respectively. We use $C_d=C_h=768$ and dropout $0.1$.

\begin{table*}[t]
\centering
\small
\setlength{\tabcolsep}{2pt}
\begin{tabular}{*{7}{>{\centering\arraybackslash}p{0.132\textwidth}}}
\hline
\noalign{\vskip 2pt}
\multicolumn{1}{c}{\raisebox{0.5\height}{Test data}} & \multicolumn{1}{c}{\raisebox{0.5\height}{Method}} & \multicolumn{1}{c}{\raisebox{0.5\height}{AP$\uparrow$}} & \multicolumn{1}{c}{\raisebox{0.5\height}{ROC AUC$\uparrow$}} & \multicolumn{1}{c}{\shortstack{Prec@Recall\\=0.85$\uparrow$}} & \multicolumn{1}{c}{\shortstack{Recall@Prec\\=0.99$\uparrow$}} & \multicolumn{1}{c}{\raisebox{0.5\height}{Time (ms)$\downarrow$}} \\
\noalign{\vskip 2pt}
\hline
DG & DG-OG & 0.956 & 0.947 & 0.910 & 0.614 & 191.5 \\
DG & DG++ & \textbf{0.981} & \textbf{0.981} & \textbf{0.982} & \underline{0.642} & \underline{118.3} \\
DG & XDG & \underline{0.978} & \underline{0.975} & \underline{0.963} & \textbf{0.702} & \textbf{34.2} \\
\hline
VisymScenes & DG-OG & 0.938 & 0.921 & 0.831 & 0.623 & 177.9 \\
VisymScenes & DG++ & \underline{0.991} & \underline{0.990} & \textbf{0.999} & \underline{0.901} & \underline{115.1} \\
VisymScenes & XDG & \textbf{0.995} & \textbf{0.994} & \underline{0.994} & \textbf{0.905} & \textbf{33.5} \\
\hline
\end{tabular}
\caption{\textbf{Pairwise visual disambiguation.} We compare DG-OG, DG++, and XDG on the DG and VisymScenes test sets. Runtime is reported as the average inference time per image pair in milliseconds. For DG-OG, runtime includes the LoFTR~\cite{sun2021loftr} matching step required as input to the classifier. XDG is more than $3\times$ faster than DG++ while maintaining comparable disambiguation performance. Best results are shown in bold, and second-best results are underlined.}
\vspace{-1 em}
\label{tab:pairwise_disambiguation}
\end{table*}

\subsection{Implementation Details}
Given two-class logits $z$, the probability of a true match is $p=\mathrm{softmax}(z)_1$, while the doppelganger (false-match) probability is $\mathrm{softmax}(z)_0$. We train XDG with focal loss~\cite{lin2017focal}, which down-weights already-easy pairs and focuses learning on ambiguous examples. For a labeled pair with ground-truth class $y \in \{0,1\}$ and predicted probability $p_y=\mathrm{softmax}(z)_y$, the loss is
\begin{equation}
    \mathcal{L}_{\mathrm{focal}}
    = - (1-p_y)^\gamma \log p_y,
\end{equation}
with $\gamma=1$ in our implementation.

We follow DG++~\cite{xiangli2025doppelgangers} and train our method using labeled image pairs from the DG dataset~\cite{cai2023doppelgangers} and VisymScenes~\cite{xiangli2025doppelgangers}. During training, we randomly flip the input order with probability $0.5$ and sample multiple resized resolutions to improve robustness to viewpoint ordering and aspect-ratio changes. We train for 10 epochs using the AdamW optimizer~\cite{loshchilov2019decoupled} with a learning rate of $1\times10^{-4}$ for all trainable parameters, including the LoRA adapters, camera tokens, fusion module, and classifier head; we set $\beta_1=0.9$, $\beta_2=0.999$, and the weight decay to $0.05$. The learning rate is linearly warmed up for one epoch, then decayed with a step schedule starting at epoch 3 using a decay factor of $0.8$ and a minimum learning rate of $1\times10^{-8}$. On 8 NVIDIA H100 GPUs, the model converges in 4 hours.

\section{Experiments}
\label{sec:experiments}

\begin{figure}[t]
\centering
\includegraphics[width=\columnwidth]{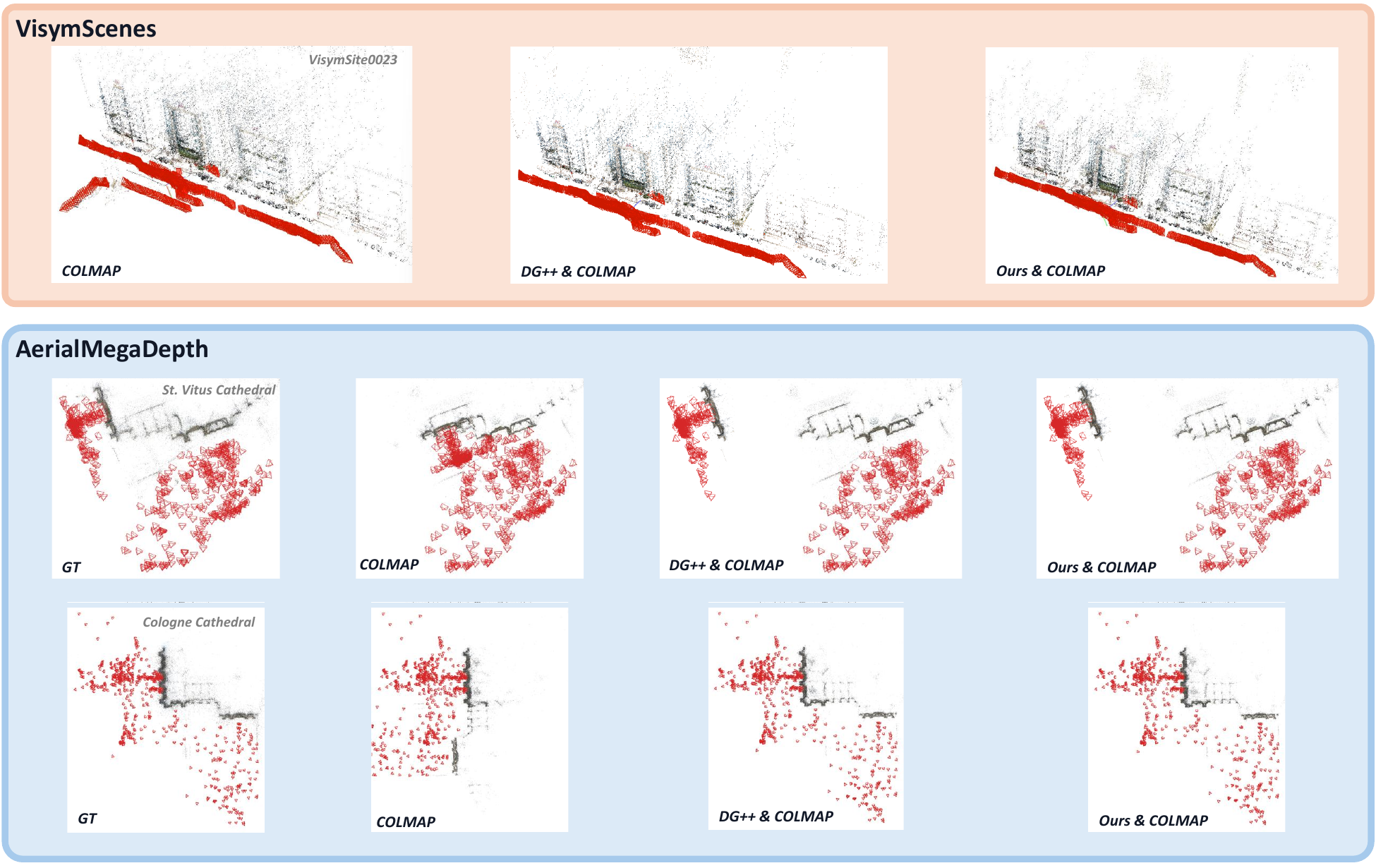}
\vspace{-1em}
\caption{\textbf{Qualitative COLMAP reconstructions.} Top: VisymSite0023 from VisymScenes reconstructed with vanilla COLMAP, DG++ filtering, and XDG filtering. Vanilla COLMAP produces a model with incorrectly registered images, while both disambiguation methods separate these cameras into a different model. Bottom: Two scenes from AerialMegaDepth. From left to right, we show the ground-truth camera layout, vanilla COLMAP, DG++ with COLMAP, and XDG with COLMAP. Without doppelganger filtering, physically distinct surfaces with similar appearance are collapsed into one, whereas DG++ and XDG successfully separate them.}
\vspace{-1em}
\label{fig:sfm_qualitative}
\end{figure}

\begin{table*}[t]
\centering
\scriptsize
\renewcommand{\arraystretch}{1}
\setlength{\tabcolsep}{2pt}
\begin{tabular}{*{4}{>{\centering\arraybackslash}p{0.096\textwidth}}@{\hspace{6pt}}*{3}{>{\centering\arraybackslash}p{0.096\textwidth}}@{\hspace{8pt}}*{2}{>{\centering\arraybackslash}p{0.096\textwidth}}}
\hline
& \multicolumn{3}{c}{\# SfM-registered images} & \multicolumn{3}{c}{Inlier ratio$\uparrow$} & \multicolumn{2}{c}{Time (h)$\downarrow$} \\
\cmidrule(lr){2-4}\cmidrule(lr){5-7}\cmidrule(lr){8-9}
Test scene & COLMAP & DG++ & XDG & COLMAP & DG++ & XDG & DG++ & XDG \\
\hline
VisymSite0010 & 1448 & 1393 & 1412 & 0.704 & 0.866 & 0.876 & 1.49 & 0.40 \\
VisymSite0023 & 654 & $571 + 83$ & $571 + 82$ & 0.863 & 0.974 & 0.977 & 0.89 & 0.24 \\
VisymSite0028 & 438 & 450 & $206 + 120 + 61$ & 0.797 & 0.856 & 0.993 & 1.14 & 0.31 \\
VisymSite0042 & 1336 & 1319 & 1306 & 0.930 & 0.956 & 0.958 & 1.01 & 0.27 \\
VisymSite0109 & 814 & $608 + 117$ & $594 + 170$ & 0.960 & 0.998 & 0.981 & 1.49 & 0.40 \\
\hline
\end{tabular}
\caption{\textbf{COLMAP reconstruction on VisymScenes.} We compare XDG against vanilla COLMAP and DG++ on five VisymScenes test scenes. We report the number of registered images, geo-alignment inlier ratio, and total visual-disambiguation time per scene. XDG achieves competitive geo-alignment while significantly reducing disambiguation time.}
\label{tab:visymsite_sfm}
\vspace{-1 em}
\end{table*}

\subsection{Pairwise Visual Disambiguation}
\label{sec:pairwise_disambiguation}

\subsubsection{Datasets and metrics.} We evaluate pairwise visual disambiguation on the DG~\cite{cai2023doppelgangers} and VisymScenes~\cite{xiangli2025doppelgangers} test sets following the testing protocol of DG++~\cite{xiangli2025doppelgangers}. We report average precision (AP), ROC AUC, precision at recall $0.85$, recall at precision $0.99$, and average inference time per image pair. AP and ROC AUC measure threshold-free ranking quality, while the fixed precision--recall operating points reflect the need to reject false edges without removing too much graph connectivity. We compare XDG with DG-OG~\cite{cai2023doppelgangers} and DG++~\cite{xiangli2025doppelgangers} in the same software environment on an NVIDIA RTX 4090. The runtime for DG-OG includes its required LoFTR~\cite{sun2021loftr} preprocessing.

\subsubsection{Results.} Quantitative results are shown in Tab.~\ref{tab:pairwise_disambiguation}. On the DG test set, XDG achieves $0.978$ AP and $0.975$ ROC AUC, approaching the performance of DG++ while substantially outperforming DG-OG. More importantly, XDG improves recall at the strict $0.99$ precision operating point from $0.642$ to $0.702$ compared with DG++. This is particularly favorable for downstream SfM, where falsely removing true matches can disconnect the match graph and harm reconstruction. High precision ensures that most retained edges are reliable, while higher recall preserves more true matches and therefore more graph connectivity. On VisymScenes, XDG slightly outperforms DG++ in AP, ROC AUC, and recall at $0.99$ precision, demonstrating that XDG achieves comparable or stronger pairwise disambiguation performance across benchmarks.

In terms of efficiency, XDG reduces the per-pair runtime of DG++ from $118.3$ ms to $34.2$ ms on DG and from $115.1$ ms to $33.5$ ms on VisymScenes. Compared to DG-OG, XDG is also substantially faster, reducing runtime from $191.5$ ms to $34.2$ ms on DG and from $177.9$ ms to $33.5$ ms on VisymScenes. Overall, XDG runs more than $3\times$ faster than DG++ and more than $5\times$ faster than DG-OG, while maintaining competitive pairwise accuracy. These results support our central claim that doppelganger disambiguation can be made significantly more efficient without sacrificing performance.

\subsection{SfM Reconstruction}
\label{sec:sfm}

XDG operates on the candidate image graph and can be seamlessly integrated into both COLMAP~\cite{schonberger2016structure} and GLUEMAP~\cite{pan2026gluemap}. We evaluate SfM performance on four datasets. VisymScenes~\cite{xiangli2025doppelgangers} and AerialMegaDepth~\cite{vuong2025aerialmegadepth} are in the training domain of both DG++ and XDG, while WRIVA~\cite{cjk5-gf33-24} and LaMAR~\cite{sarlin2022lamar} test out-of-domain generalization. Because VisymScenes has noisy geotags and lacks accurate scene-wide camera ground truth, DG++ evaluates its reconstruction quality using the inlier ratio. While this metric reflects geometric consistency after reconstruction, it provides only an indirect measure of camera accuracy. In contrast, AerialMegaDepth, WRIVA, and LaMAR provide camera calibrations, geospatial references, or laser-scan-aligned poses, allowing pose-based evaluation of the reconstructions.

\subsubsection{Datasets}

VisymScenes and AerialMegaDepth are reconstructed with COLMAP, while WRIVA and LaMAR are reconstructed with GLUEMAP. We present the training-domain datasets first, followed by the out-of-domain datasets.

\textbf{VisymScenes.} Introduced by DG++~\cite{xiangli2025doppelgangers}, VisymScenes contains 258K images with GPS/IMU metadata collected at 149 sites across 42 cities and 15 countries. It covers landmarks as well as everyday residential, rural, suburban, and business environments, many of which contain repeated structures and visually similar but physically distinct surfaces. We use the same 5 test scenes evaluated by DG++.

\textbf{AerialMegaDepth.} AerialMegaDepth~\cite{vuong2025aerialmegadepth} contains 132K images across 137 scenes and co-registers real images from MegaDepth~\cite{li2018megadepth} with pseudo-synthetic aerial and ground-level views rendered from geospatial 3D meshes with known camera poses. The rendered views provide useful pose references and help reduce ambiguity caused by visually similar or duplicated structures, making the dataset suitable for evaluating visual disambiguation. However, because the real images are registered to geospatial meshes rather than captured with ground-truth camera calibration, their poses should be treated as approximate rather than fully reliable. We therefore use AerialMegaDepth primarily as a large-scale, challenging benchmark for reconstruction consistency, evaluating on 8 scenes with duplicated structures and using only the real images as reconstruction inputs.

\textbf{WRIVA.} WRIVA~\cite{cjk5-gf33-24} contains calibrated imagery captured from heterogeneous viewpoints and altitudes. Camera locations are georeferenced using RTK-corrected GPS with centimeter-level accuracy, providing reliable camera-position ground truth for evaluating large-scale reconstruction and alignment. We use 34 sequences for evaluation.

\textbf{LaMAR.} LaMAR~\cite{sarlin2022lamar} contains large indoor and outdoor scenes captured along unconstrained AR-device trajectories, with accurate reference poses obtained by registering the trajectories to laser scans. We evaluate on the same benchmark splits used by GLUEMAP~\cite{pan2026gluemap}.

\begin{table}[t]
\centering
\small
\setlength{\tabcolsep}{3pt}
\begin{tabular}{@{}lccccc@{}}
\hline
\noalign{\vskip 2pt}
\multicolumn{1}{c}{\multirow{2}{*}{\raisebox{-0.45ex}{Method}}} & \multicolumn{4}{c}{Pose AUC (\%)$\uparrow$} & \multicolumn{1}{c}{\multirow{2}{*}{\raisebox{-0.45ex}{Total time (h)$\downarrow$}}} \\
\cmidrule(lr){2-5}
& @$3^{\circ}$ & @$5^{\circ}$ & @$10^{\circ}$ & @$30^{\circ}$ & \\
\noalign{\vskip 2pt}
\hline
No disamb. & 49.70 & 55.79 & 61.18 & 65.56 & -- \\
DG++ & 57.26 & 64.45 & 70.75 & 75.63 & 5.28 \\
XDG & 57.57 & 64.68 & 70.98 & 75.94 & 1.20 \\
\hline
\end{tabular}
\caption{\textbf{COLMAP reconstruction on AerialMegaDepth.} Pose AUC is averaged over 8 scenes using only real images as reconstruction inputs, while time is the total visual-disambiguation runtime over all scenes. DG++ and XDG both improve reconstruction accuracy over no doppelganger filtering and achieve nearly identical accuracy, while XDG reduces total disambiguation time from $5.28$ to $1.20$ hours, saving $4.08$ hours ($4.40\times$).}
\label{tab:aerialmegadepth_colmap}
\end{table}

\begin{table}[t]
\centering
\small
\setlength{\tabcolsep}{3pt}
\begin{tabular}{@{}lccccc@{}}
\hline
\noalign{\vskip 2pt}
\multicolumn{1}{c}{\multirow{2}{*}{\raisebox{-0.45ex}{Method}}} & \multicolumn{4}{c}{Pose AUC (\%)$\uparrow$} & \multicolumn{1}{c}{\multirow{2}{*}{\raisebox{-0.45ex}{Total time (h)$\downarrow$}}} \\
\cmidrule(lr){2-5}
& @$3^{\circ}$ & @$5^{\circ}$ & @$10^{\circ}$ & @$30^{\circ}$ & \\
\noalign{\vskip 2pt}
\hline
No disamb. & 13.16 & 19.49 & 29.46 & 47.52 & -- \\
DG++ & 22.06 & 31.95 & 45.14 & 62.22 & 9.86\\
XDG & 21.24 & 31.33 & 45.43 & 63.73 & 3.06\\
\hline
\end{tabular}
\caption{\textbf{GLUEMAP reconstruction on WRIVA.} Pose AUC is averaged across 34 sequences, while time is the total visual-disambiguation runtime over the complete benchmark. Higher AUC is better. Both DG++ and XDG substantially improve upon reconstruction without doppelganger detection, with DG++ slightly outperforming XDG in the lowest error regime at the cost of more than triple the runtime. \vspace{-1 em}}
\label{tab:wriva_auc}
\end{table}

\begin{table}[t]
\centering
\small
\setlength{\tabcolsep}{1.5pt}
\begin{tabular}{@{}llccccc@{}}
\hline
\noalign{\vskip 2pt}
\multicolumn{1}{c}{\multirow{2}{*}{\raisebox{-0.45ex}{Scene (\# images)}}} & \multicolumn{1}{c}{\multirow{2}{*}{\raisebox{-0.45ex}{Method}}} & \multicolumn{4}{c}{Pose AUC (\%)$\uparrow$} & \multicolumn{1}{c}{\multirow{2}{*}{\raisebox{-0.45ex}{Total time (h)$\downarrow$}}} \\
\cmidrule(lr){3-6}
& & @$3^{\circ}$ & @$5^{\circ}$ & @$10^{\circ}$ & @$30^{\circ}$ & \\
\noalign{\vskip 2pt}
\hline
CAB ($6{,}587$) & DG++ & 3.70 & 8.90 & 27.57 & 65.77 & 14.18 \\
    & XDG  & 3.70 & 8.11 & 20.00 & 51.44 & 4.29 \\
\hline
HGE ($7{,}553$) & DG++ & 36.52 & 56.54 & 76.91 & 91.80 & 15.73 \\
    & XDG  & 38.16 & 55.87 & 76.18 & 90.73 & 4.88 \\
\hline
LIN ($9{,}319$) & DG++ & 43.58 & 60.48 & 74.45 & 85.37 & 20.02 \\
    & XDG  & 43.79 & 60.91 & 74.81 & 85.57 & 6.11 \\
\hline
\end{tabular}
\caption{\textbf{GLUEMAP reconstruction on LaMAR.} We replace GLUEMAP's DG++ disambiguator with XDG and keep all other stages fixed. DG++ generalizes slightly better on the indoor CAB scene, whereas XDG performs comparably on the outdoor HGE and LIN scenes while reducing total disambiguation time by $3.27\times$.}
\label{tab:lamar_GLUEMAP}
\end{table}

\subsubsection{Implementation Details and Metrics}

For each reconstruction experiment, we run visual disambiguation before reconstruction and keep all remaining stages fixed.

\textbf{COLMAP.} We follow the DG++ protocol and use COLMAP's vocabulary-tree matcher~\cite{schoenberger2016vote} with default settings and confidence threshold $\tau=0.8$.

\textbf{GLUEMAP.} We replace GLUEMAP's default DG++ two-view stage with XDG while keeping retrieval, feed-forward local reconstruction, global mapping, and refinement unchanged. We use the benchmark configuration reported by GLUEMAP for LaMAR, and default settings for WRIVA.

\textbf{Metrics.} On VisymScenes, we report the number of registered images and the geo-alignment inlier ratio used by DG++~\cite{xiangli2025doppelgangers}, since the GPS annotations are noisy. Specifically, reconstructed camera centers are aligned to GPS locations with RANSAC, and the inlier ratio measures the fraction of registered images that are consistent with the estimated alignment. For reconstructions containing multiple disconnected components, we report the weighted inlier ratio
\begin{equation}
\mathrm{IR}_{\mathrm{weighted}} =
\frac{\sum_k n_k \mathrm{IR}_k}{\sum_k n_k},
\end{equation}
where $n_k$ and $\mathrm{IR}_k$ denote the number of registered images and the inlier ratio of component $k$, respectively. On AerialMegaDepth, WRIVA, and LaMAR, we report AUC@$X^{\circ}$, defined as the normalized area under the recall curve of pairwise relative-pose error up to the angular threshold $X^{\circ}$. For each image pair, the pose error is computed as the maximum of the relative rotation error and the translation-direction error. Image pairs for which one or both cameras are not registered are counted as failures and therefore have an error of infinity. Tight thresholds emphasize pose accuracy, while looser thresholds also reflect reconstruction completeness. We use $X\in\{3,5,10,30\}$. All reported times correspond to visual-disambiguation runtime and are measured in hours. For AerialMegaDepth, WRIVA, and LaMAR, the tables report total inference time summed over every scene or sequence in the respective benchmark rather than an average per scene. Experiments on these three datasets are run in the same software environment on a cluster with 8 NVIDIA A100 40~GB GPUs.

\begin{figure}[t]
\centering
\includegraphics[width=\linewidth]{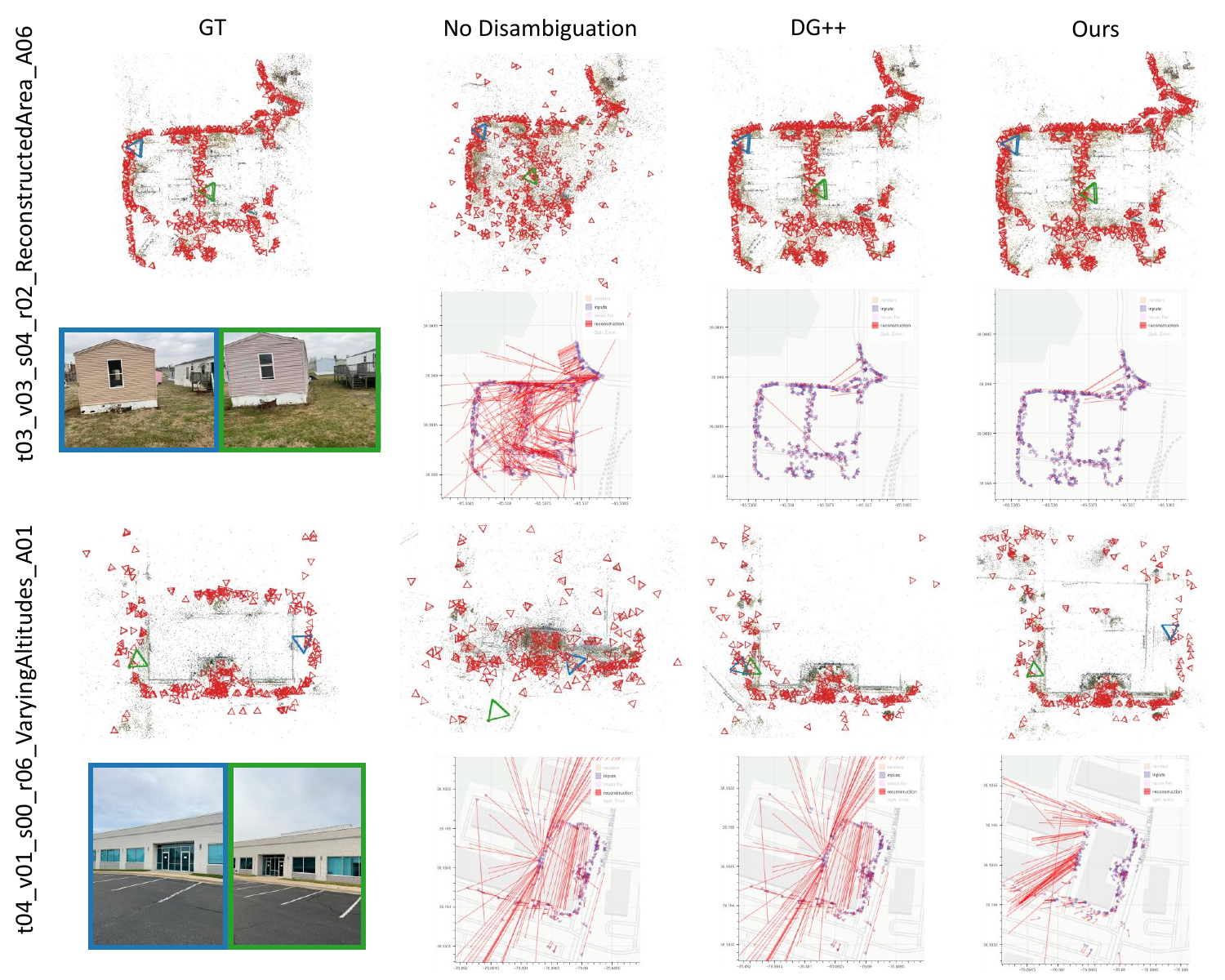}
\caption{\textbf{Qualitative GLUEMAP reconstruction.} We show two WRIVA sequences, with the top two rows corresponding to one sequence and the bottom two rows to another. For each sequence, the first row shows the ground-truth camera layout and reconstructions without disambiguation, with DG++, and with XDG; blue and green cameras indicate the ambiguous image pair shown below. The second row shows the image pair and the corresponding alignment to ground truth.}
\label{fig:wriva_qualitative}
\end{figure}

\subsubsection{Results}
\textbf{VisymScenes.} Tab.~\ref{tab:visymsite_sfm} reports the largest COLMAP models after disambiguation. The ``$+$'' symbol denotes split reconstruction components. XDG achieves reconstruction quality comparable to DG++ while substantially reducing the required disambiguation time. Across the five VisymScenes scenes, XDG registers a similar number of images and achieves similar inlier ratios. Qualitative results in Fig.~\ref{fig:sfm_qualitative} further show that XDG removes incorrectly registered cameras and produces camera layouts visually comparable to DG++ on VisymSite0023. Averaged across scenes, XDG reduces disambiguation time from $1.20$ hours to $0.32$ hours, corresponding to a $3.7\times$ speedup.

\textbf{AerialMegaDepth.} Tab.~\ref{tab:aerialmegadepth_colmap} reports COLMAP reconstruction accuracy across 8 scenes. Both XDG and DG++ substantially improve pose AUC over reconstruction without disambiguation and perform nearly identically. As shown qualitatively in Fig.~\ref{fig:sfm_qualitative}, doppelganger filtering prevents visually similar but physically distinct structures from being incorrectly collapsed, and XDG preserves the reconstruction quality of DG++. The main advantage of XDG is efficiency: under the same inference setup, XDG reduces total visual-disambiguation time over all 8 scenes from $5.28$ to $1.20$ hours, saving $4.08$ hours.

\begin{table}[t]
\centering
\scriptsize
\setlength{\tabcolsep}{2pt}
\begin{tabular}{@{}>{\raggedright\arraybackslash}m{0.3\columnwidth}
                *{3}{>{\centering\arraybackslash}m{0.2\columnwidth}}@{}}
\hline
\noalign{\vskip 2pt}
\multicolumn{1}{c}{Variant} &
\multicolumn{1}{c}{AP$\uparrow$} &
\multicolumn{1}{c}{ROC AUC$\uparrow$} &
\multicolumn{1}{c}{Time (ms)$\downarrow$} \\
\noalign{\vskip 2pt}
\hline
\multicolumn{4}{@{}l}{\textit{DG}} \\
DA3 + transformer & 0.9617 & 0.9596 & 111.9 \\
\quad + feature aggregation & 0.9591 & 0.9557 & 55.4 \\
\quad + MLP (dense tokens) & 0.9586 & 0.9551 & 38.9 \\
\quad + camera tokens & 0.9540 & 0.9511 & \textbf{32.9} \\
\quad + LoRA (XDG) & \textbf{0.9780} & \textbf{0.9750} & 34.2 \\
\hline
\multicolumn{4}{@{}l}{\textit{VisymScenes}} \\
DA3 + transformer & 0.9919 & 0.9916 & 110.7 \\
\quad + feature aggregation & 0.9922 & 0.9908 & 54.6 \\
\quad + MLP (dense tokens) & 0.9912 & 0.9904 & 38.8 \\
\quad + camera tokens & 0.9905 & 0.9896 & \textbf{32.4} \\
\quad + LoRA (XDG) & \textbf{0.9950} & \textbf{0.9940} & 33.5 \\
\hline
\end{tabular}
\caption{\textbf{Cumulative ablation of XDG.} Starting from a frozen DA3 backbone with a dense-token transformer classifier following DG++, each row adds the listed change. The best accuracy and lowest latency in each dataset block are highlighted. The full ablation table is provided in the supplementary material.}
\label{tab:xdg_ablation}
\end{table}

\textbf{WRIVA.} Tab.~\ref{tab:wriva_auc} reports out-of-domain pose accuracy averaged over 34 sequences, and Fig.~\ref{fig:wriva_qualitative} shows qualitative results on two representative sequences. Both DG++ and XDG substantially outperform reconstruction without disambiguation. DG++ performs best at the tightest thresholds, suggesting stronger generalization to these unseen scenes, while XDG achieves slightly higher AUC@$10^{\circ}$ and AUC@$30^{\circ}$ and preserves most of the reconstruction completeness with a substantially lower inference cost. In total, XDG reduces visual-disambiguation time over the 34 sequences from $9.86$ to $3.06$ hours, saving $6.80$ hours. Qualitatively, both DG++ and XDG remove most visually plausible false matches that would otherwise distort the reconstruction, producing camera trajectories that align closely with the ground truth.

\textbf{LaMAR.} Tab.~\ref{tab:lamar_GLUEMAP} reports results on LaMAR~\cite{sarlin2022lamar}. On CAB, an indoor sequence, DG++ generalizes better: it ties XDG at AUC@$3^{\circ}$ and performs better at the looser thresholds. On the outdoor HGE and LIN scenes, the two methods are comparable. Across all three scenes, XDG reduces total visual-disambiguation time from $49.93$ to $15.28$ hours, saving $34.65$ hours at a $3.27\times$ speedup. The scaling benefit is already substantial on individual scenes: HGE contains $7{,}553$ images and saves $10.85$ hours, while LIN contains $9{,}319$ images and saves $13.91$ hours. These results show that DG++ already works well across many cases, but its runtime is a bottleneck for large image collections; XDG directly alleviates this bottleneck.

\subsection{Ablation}
\label{sec:ablation}

We ablate the main design choices of XDG on pairwise classification in Tab.~\ref{tab:xdg_ablation}. The study is cumulative: each row adds one modification to the configuration above it. We start from a \textbf{DA3--transformer baseline}, which replaces the MASt3R~\cite{leroy2024grounding} backbone in DG++~\cite{xiangli2025doppelgangers} with frozen DA3~\cite{lin2025depth}. We then introduce \textbf{feature aggregation}, which aligns and fuses features from the two input orders, removing the need for duplicated classification heads. Next, we replace the transformer classifier with a lightweight \textbf{MLP head} to further reduce inference cost. The \textbf{camera-token} variant further compresses the representation by replacing dense patch tokens with DA3 camera tokens for classification. Finally, we add \textbf{LoRA} adapters to adapt the frozen DA3 backbone to the disambiguation task. The last row corresponds to the full XDG model reported in Tab.~\ref{tab:pairwise_disambiguation}.

Feature aggregation substantially reduces latency by removing redundant classification over the two input orders, while the MLP head further lowers the computational cost with little change in overall accuracy. Using DA3 camera tokens as the pair-level representation gives the fastest model, indicating that these tokens provide a compact and effective summary for pairwise disambiguation. However, this compression also removes some discriminative signal present in the dense tokens, leading to a small accuracy drop compared with the preceding variants.

LoRA fine-tuning recovers this lost discriminative signal with little additional runtime. Compared with the camera-token variant, the full XDG model consistently improves AP and ROC AUC on both test sets. The final model remains more than three times faster than the baseline while achieving the best accuracy. Overall, the ablation separates the roles of the main proposed components: feature aggregation, the MLP head, and camera-token classification provide the acceleration, while LoRA fine-tuning improves the accuracy needed for robust SfM disambiguation.

\section{Conclusion}
\label{sec:conclusion}
We introduced XDG, an efficient visual-disambiguation model for removing doppelganger edges in large-scale 3D reconstruction. The state-of-the-art method DG++ already resolves many difficult ambiguities effectively, but its runtime becomes a bottleneck when scaled to large image collections with thousands of images. By adapting a 3D foundation model with LoRA and replacing heavy transformer classifiers with lightweight token aggregation and an MLP head, XDG achieves performance comparable to DG++ while reducing disambiguation time by more than $3\times$ across pairwise and downstream reconstruction tasks. This speedup saves $34.65$ hours in total over the LaMAR benchmark and more than $6$ hours on the WRIVA dataset. Our extensive experiments show that XDG can make visual disambiguation significantly more efficient while preserving reconstruction accuracy.

\section*{Acknowledgments}
This material is based upon work supported by the Intelligence Advanced Research Projects Activity under prime Contract No. 140D0423C0034. The U.S. Government is authorized to reproduce and distribute reprints for governmental purposes notwithstanding any copyright annotation thereon. Disclaimer: The views and conclusions contained herein are those of the authors and should not be interpreted as necessarily representing the official policies or endorsements, either expressed or implied, of IARPA, DOI/IBC, or the U.S. Government.

{
    \small
    \bibliographystyle{ieeenat_fullname}
    \bibliography{main}
}

\clearpage
\appendix
\twocolumn[{
    \vspace*{0.375in}
    \centering
    {\Large\bfseries Supplementary Material\par}
    \vspace{1em}
}]

\section{Additional Reconstruction Results}

\subsection{WRIVA}

\paragraph{COLMAP.}
We additionally evaluate COLMAP~\cite{schonberger2016structure} on the 34 WRIVA~\cite{cjk5-gf33-24} sequences in Tab.~\ref{tab:wriva_colmap}. In contrast to the GLUEMAP~\cite{pan2026gluemap} results in the main paper, both DG++~\cite{xiangli2025doppelgangers} and XDG reduce pose AUC relative to vanilla COLMAP.

\begin{table}[H]
\centering
\small
\setlength{\tabcolsep}{3.5pt}
\begin{tabular}{@{}lcccc@{}}
\hline
\noalign{\vskip 2pt}
Method &
\multicolumn{4}{c}{Pose AUC (\%)$\uparrow$} \\
\cmidrule(lr){2-5}
& @$3^{\circ}$ & @$5^{\circ}$ & @$10^{\circ}$ & @$30^{\circ}$ \\
\noalign{\vskip 2pt}
\hline
No disamb. & 16.60 & 23.33 & 30.94 & 38.66 \\
DG++ & 14.22 & 19.68 & 25.75 & 31.70 \\
XDG & 14.16 & 19.45 & 25.53 & 31.46 \\
\hline
\end{tabular}
\caption{\textbf{COLMAP reconstruction on WRIVA.} Pose AUC is averaged across 34 sequences. Higher is better. On this challenging dataset, filtering candidate edges with either DG++ or XDG reduces reconstruction accuracy.}
\label{tab:wriva_colmap}
\end{table}

WRIVA contains substantial variation in viewpoint, altitude, and appearance. Under these conditions, COLMAP's conventional feature matching may fail to establish a sufficiently complete set of reliable correspondences. Visual disambiguation then prunes an already sparse match graph, potentially leaving too few tracks for accurate and complete reconstruction. These results suggest that correspondence formation, rather than false-edge removal alone, is the primary bottleneck for the COLMAP pipeline on WRIVA and motivate our use of the more robust GLUEMAP pipeline for the main WRIVA experiments.

\subsection{AerialMegaDepth}

\paragraph{GLUEMAP.}
We also evaluate GLUEMAP reconstruction on the eight AerialMegaDepth~\cite{vuong2025aerialmegadepth} scenes in Tab.~\ref{tab:aerialmegadepth_gluemap}. Both DG++ and XDG substantially improve pose AUC over reconstruction without visual disambiguation, and the two methods perform comparably across all thresholds.

\begin{table}[H]
\centering
\small
\setlength{\tabcolsep}{3.5pt}
\begin{tabular}{@{}lcccc@{}}
\hline
\noalign{\vskip 2pt}
Method &
\multicolumn{4}{c}{Pose AUC (\%)$\uparrow$} \\
\cmidrule(lr){2-5}
& @$3^{\circ}$ & @$5^{\circ}$ & @$10^{\circ}$ & @$30^{\circ}$ \\
\noalign{\vskip 2pt}
\hline
No disamb. & 19.40 & 28.21 & 41.12 & 59.19 \\
DG++ & 34.09 & 46.05 & 60.94 & 77.81 \\
XDG & 35.07 & 46.98 & 60.88 & 77.04 \\
\hline
\end{tabular}
\caption{\textbf{GLUEMAP reconstruction on AerialMegaDepth.} Pose AUC is averaged across eight scenes using only real images as reconstruction inputs. Higher is better. Both visual-disambiguation methods substantially improve reconstruction accuracy and perform comparably.}
\label{tab:aerialmegadepth_gluemap}
\end{table}

These GLUEMAP results are lower than the COLMAP results reported in the main paper. Some real AerialMegaDepth images are difficult to register or reconstruct without the rendered images provided in the original dataset. GLUEMAP attempts to incorporate all input images, including these difficult views, which can introduce inaccurate camera estimates and reduce the aggregate pose AUC. Nevertheless, the clear improvement from both DG++ and XDG shows that visual disambiguation remains beneficial within this reconstruction setting.

\begin{table*}[ht]
\centering
\small
\setlength{\tabcolsep}{4pt}
\begin{tabular}{>{\raggedright\arraybackslash}m{0.34\textwidth}
                *{5}{>{\centering\arraybackslash}m{0.105\textwidth}}}
\hline
\noalign{\vskip 2pt}
\multicolumn{1}{c}{\raisebox{0.5\height}{Variant}} &
\multicolumn{1}{c}{\raisebox{0.5\height}{AP$\uparrow$}} &
\multicolumn{1}{c}{\raisebox{0.5\height}{ROC AUC$\uparrow$}} &
\multicolumn{1}{c}{\shortstack{Prec@Recall\\=0.85$\uparrow$}} &
\multicolumn{1}{c}{\shortstack{Recall@Prec\\=0.99$\uparrow$}} &
\multicolumn{1}{c}{\shortstack{Time/pair\\(ms)$\downarrow$}} \\
\noalign{\vskip 2pt}
\hline
\multicolumn{6}{l}{\textit{DG}} \\
DA3 + transformer & 0.9617 & 0.9596 & 0.9214 & 0.5195 & 111.9 \\
\quad + feature aggregation & 0.9591 & 0.9557 & 0.9164 & 0.5427 & 55.4 \\
\quad + MLP (dense tokens) & 0.9586 & 0.9551 & 0.9184 & 0.4959 & 38.9 \\
\quad + camera tokens & 0.9540 & 0.9511 & 0.8992 & 0.5191 & \textbf{32.9} \\
\quad + LoRA (XDG) & \textbf{0.9780} & \textbf{0.9750} & \textbf{0.9630} & \textbf{0.7020} & 34.2 \\
\hline
\multicolumn{6}{l}{\textit{VisymScenes}} \\
DA3 + transformer & 0.9919 & 0.9916 & 0.9877 & 0.8365 & 110.7 \\
\quad + feature aggregation & 0.9922 & 0.9908 & \textbf{0.9963} & 0.9019 & 54.6 \\
\quad + MLP (dense tokens) & 0.9912 & 0.9904 & 0.9884 & 0.8434 & 38.8 \\
\quad + camera tokens & 0.9905 & 0.9896 & 0.9928 & 0.8855 & \textbf{32.4} \\
\quad + LoRA (XDG) & \textbf{0.9950} & \textbf{0.9940} & 0.9940 & \textbf{0.9050} & 33.5 \\
\hline
\end{tabular}
\caption{\textbf{Full cumulative ablation of XDG.} This table expands the ablation study in the main paper with fixed precision--recall operating points. Each row adds the listed modification. The best value for each metric and the lowest latency in each dataset block are highlighted.}
\label{tab:xdg_ablation_full}
\end{table*}

\begin{table*}[ht]
\centering
\small
\setlength{\tabcolsep}{4pt}
\begin{tabular}{>{\raggedright\arraybackslash}m{0.34\textwidth}
                *{5}{>{\centering\arraybackslash}m{0.105\textwidth}}}
\hline
\noalign{\vskip 2pt}
\multicolumn{1}{c}{\raisebox{0.5\height}{Variant}} &
\multicolumn{1}{c}{\raisebox{0.5\height}{AP$\uparrow$}} &
\multicolumn{1}{c}{\raisebox{0.5\height}{ROC AUC$\uparrow$}} &
\multicolumn{1}{c}{\shortstack{Prec@Recall\\=0.85$\uparrow$}} &
\multicolumn{1}{c}{\shortstack{Recall@Prec\\=0.99$\uparrow$}} &
\multicolumn{1}{c}{\shortstack{Time/pair\\(ms)$\downarrow$}} \\
\noalign{\vskip 2pt}
\hline
\multicolumn{6}{l}{\textit{DG}} \\
XDG (DA3-Base) & 0.9780 & 0.9750 & 0.9630 & 0.7020 & 34.2 \\
\multicolumn{6}{l}{\quad\textit{Backbone size}} \\
DA3-Small & 0.9574 & 0.9549 & 0.9163 & 0.4277 & \textbf{16.3} \\
DA3-Large & \textbf{0.9791} & \textbf{0.9762} & 0.9635 & 0.7110 & 83.2 \\
\multicolumn{6}{l}{\quad\textit{Training strategy}} \\
No random order flipping & 0.9754 & 0.9715 & 0.9602 & \textbf{0.7535} & 32.7 \\
No multi-resolution training & 0.9750 & 0.9712 & \textbf{0.9663} & 0.7312 & 32.8 \\
\multicolumn{6}{l}{\quad\textit{Order robustness}} \\
Single-order inference & 0.9751 & 0.9721 & 0.9535 & 0.6990 & 19.7 \\
\hline
\multicolumn{6}{l}{\textit{VisymScenes}} \\
XDG (DA3-Base) & \textbf{0.9950} & 0.9940 & 0.9940 & 0.9050 & 33.5 \\
\multicolumn{6}{l}{\quad\textit{Backbone size}} \\
DA3-Small & 0.9908 & 0.9894 & 0.9927 & 0.8667 & \textbf{15.5} \\
DA3-Large & 0.9948 & \textbf{0.9952} & 0.9971 & \textbf{0.9352} & 83.3 \\
\multicolumn{6}{l}{\quad\textit{Training strategy}} \\
No random order flipping & 0.9934 & 0.9928 & 0.9927 & 0.8862 & 32.8 \\
No multi-resolution training & 0.9912 & 0.9897 & 0.9905 & 0.8522 & 32.8 \\
\multicolumn{6}{l}{\quad\textit{Order robustness}} \\
Single-order inference & 0.9934 & 0.9925 & \textbf{0.9985} & 0.9157 & 19.6 \\
\hline
\end{tabular}
\caption{\textbf{Additional ablations of XDG.} XDG uses DA3-Base, random order flipping, multi-resolution training, and both input orders. Each other row changes only the named factor. The best value for each metric and the lowest latency in each dataset block are highlighted.}
\vspace{-0.5 em}
\label{tab:xdg_ablation_additional}
\end{table*}

\section{More Ablation Results}

Tab.~\ref{tab:xdg_ablation_full} expands the cumulative ablation study from the main paper with the fixed precision--recall operating points omitted there for space. The ablations are evaluated on DG~\cite{cai2023doppelgangers} and VisymScenes~\cite{xiangli2025doppelgangers}. Starting from a frozen DA3~\cite{lin2025depth} backbone with a dense-token transformer classifier, each row adds the listed modification. Feature aggregation, the MLP head, and camera-token classification reduce inference cost, while LoRA~\cite{hu2022lora} adapts the backbone to recover task accuracy.

\begin{figure*}[h]
\centering
\begin{subfigure}[t]{0.46\textwidth}
    \centering
    \includegraphics[width=\linewidth]{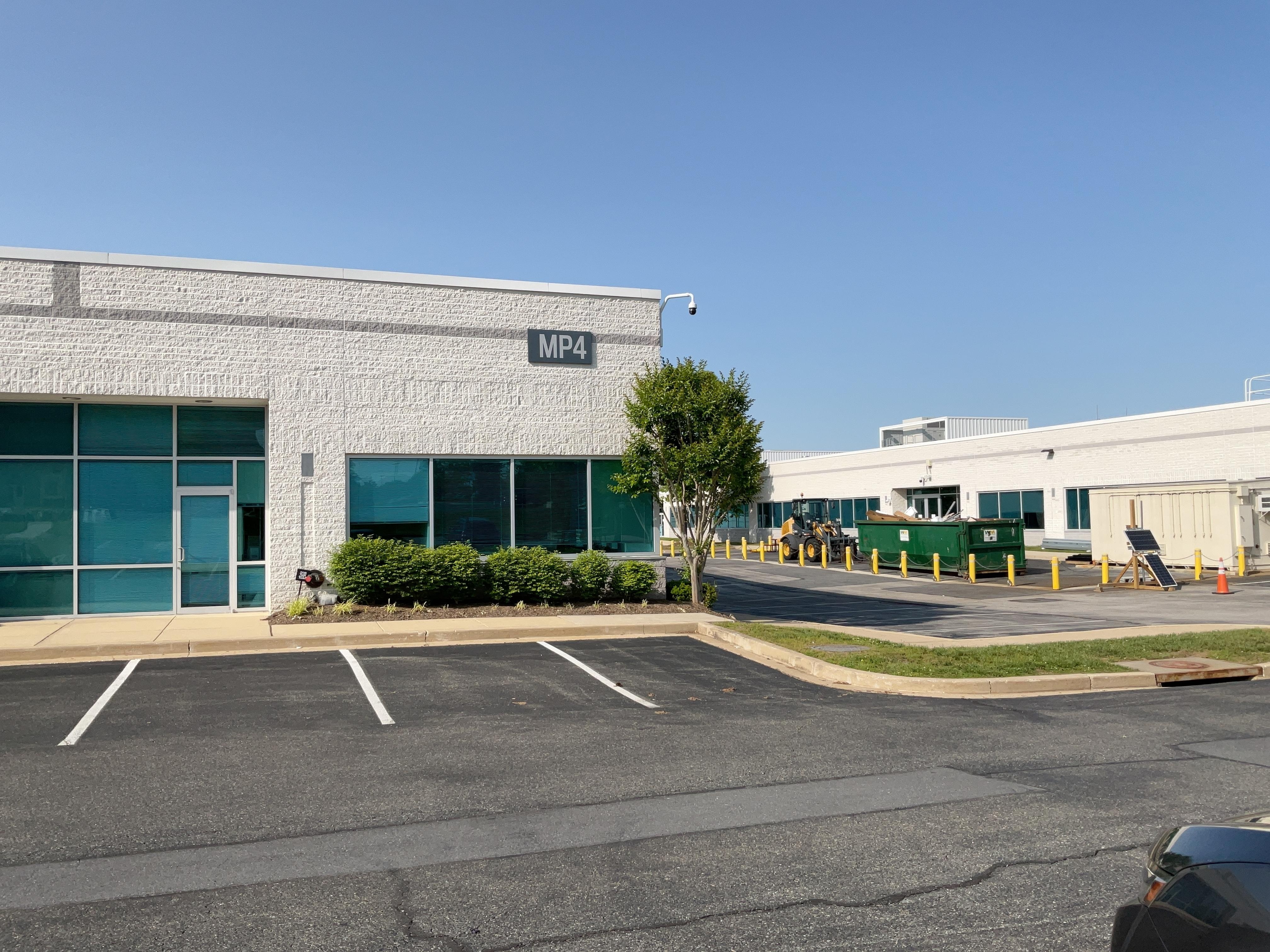}
    \caption{Image A}
\end{subfigure}
\hfill
\begin{subfigure}[t]{0.46\textwidth}
    \centering
    \includegraphics[width=\linewidth]{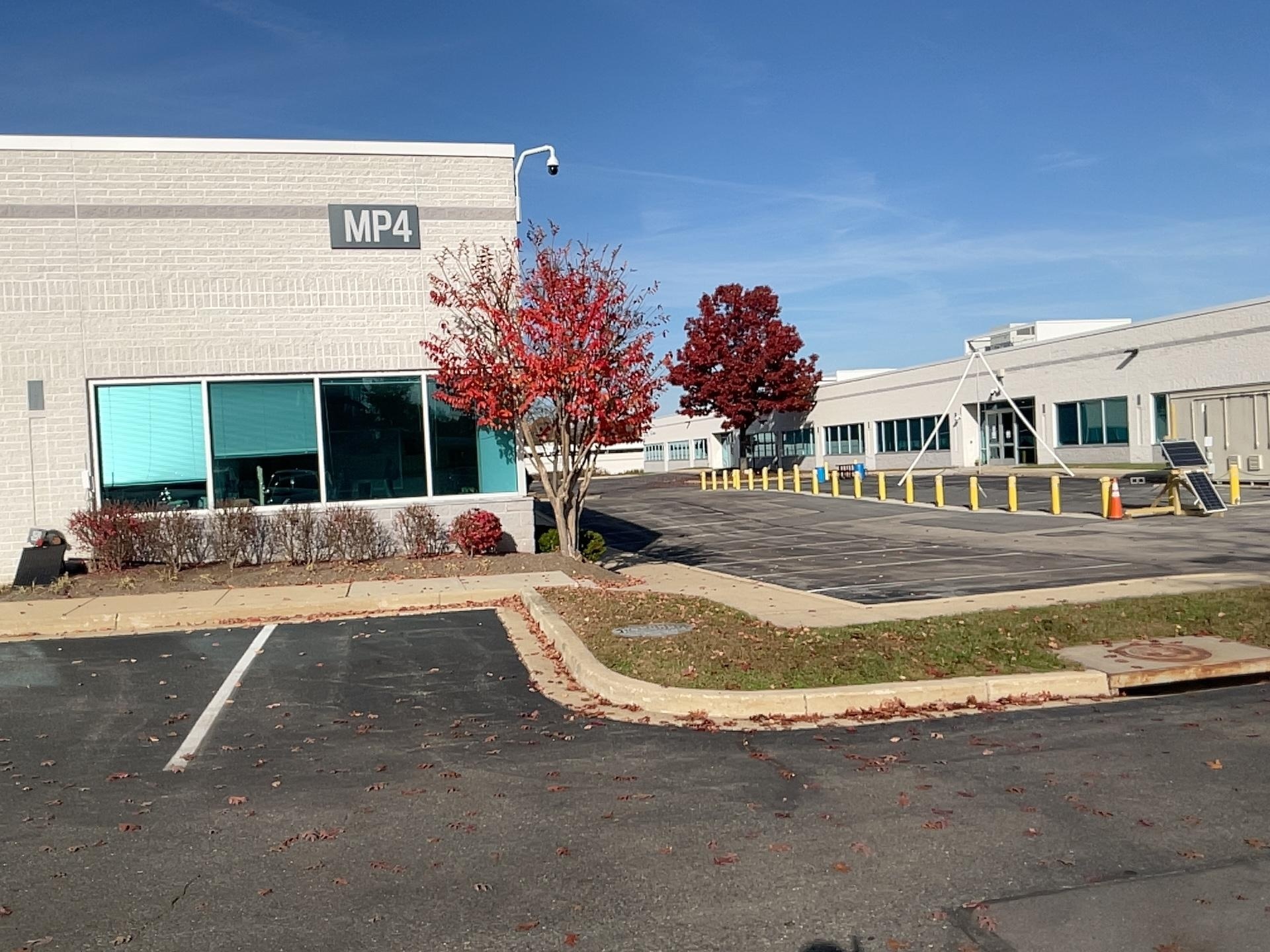}
    \caption{Image B}
\end{subfigure}
\vspace{4pt}

\small
\setlength{\tabcolsep}{8pt}
\begin{tabular}{lcccc}
\hline
\noalign{\vskip 2pt}
Model &
$p(\mathrm{match}\mid A{\rightarrow}B)$ &
$p(\mathrm{match}\mid B{\rightarrow}A)$ &
$|\Delta|$ &
Decision at $\tau=0.8$ \\
\noalign{\vskip 2pt}
\hline
XDG & 0.9727 & 0.9727 & 0.0000 & accept / accept \\
Single-order inference & 0.7734 & 0.8906 & 0.1172 & \textbf{reject} / accept \\
\hline
\end{tabular}
\caption{\textbf{Effect of bidirectional order-robust processing.} We evaluate the same image pair in both input orders. XDG aggregates features from both orders and produces the same matching confidence after swapping the inputs. The single-order model is sensitive to the input order: with a decision threshold of $\tau=0.8$, it rejects the $A{\rightarrow}B$ ordering but accepts $B{\rightarrow}A$.}
\label{fig:order_robustness}
\end{figure*}

\begin{table*}[t]
\centering
\scriptsize
\setlength{\tabcolsep}{1.5pt}
\begin{tabular}{@{}>{\raggedright\arraybackslash}m{0.175\textwidth}
                >{\centering\arraybackslash}m{0.055\textwidth}
                *{3}{>{\centering\arraybackslash}m{0.185\textwidth}}
                *{2}{>{\centering\arraybackslash}m{0.065\textwidth}}@{}}
\hline
\noalign{\vskip 2pt}
Scene & \# Images & No disamb. & DG++ & XDG &
\multicolumn{2}{c}{Time (min)$\downarrow$} \\
\cmidrule(lr){6-7}
& & & & & DG++ & XDG \\
\noalign{\vskip 2pt}
\hline
Adler Planetarium & 115 & $79.64/86.02/91.70/96.90$ & $79.04/85.68/91.53/95.89$ & $79.15/85.73/91.95/97.14$ & 10.74 & 2.37 \\
Belfry of Bruges & 348 & $39.79/49.92/59.49/67.33$ & $30.31/37.78/45.00/50.99$ & $30.54/38.03/45.22/51.22$ & 22.76 & 4.99 \\
Brussels Town Hall & 630 & $66.24/77.23/87.05/94.77$ & $66.99/77.62/87.12/94.59$ & $66.85/77.63/87.25/94.83$ & 51.55 & 10.87 \\
Cologne Cathedral & 580 & $50.19/53.70/56.62/58.84$ & $81.07/87.87/93.26/97.26$ & $80.75/87.67/93.15/97.22$ & 50.50 & 10.90 \\
Florence Baptistery & 691 & $55.31/60.91/65.41/68.61$ & $59.79/66.14/71.33/75.03$ & $61.40/67.33/72.09/75.47$ & 62.95 & 13.27 \\
The Louvre Museum & 297 & $20.85/24.97/28.53/31.74$ & $55.16/66.85/77.06/84.96$ & $55.92/67.29/77.39/85.19$ & 21.93 & 4.66 \\
Ponte di Rialto & 369 & $44.92/48.75/52.23/55.10$ & $44.89/48.73/52.23/55.12$ & $45.21/49.00/52.47/55.42$ & 42.83 & 10.24 \\
St.\ Vitus Cathedral & 500 & $40.67/44.79/48.41/51.17$ & $40.87/44.93/48.48/51.20$ & $40.76/44.80/48.34/51.05$ & 54.04 & 14.90 \\
\hline
\end{tabular}
\caption{\textbf{Per-scene COLMAP reconstruction on AerialMegaDepth.} The image count denotes the number of real input images for each scene. Each AUC cell reports pose AUC (\%) as the tuple AUC@$3^{\circ}$ / @$5^{\circ}$ / @$10^{\circ}$ / @$30^{\circ}$. Time is the visual-disambiguation runtime for each scene. Results cover all eight evaluation scenes.}
\vspace{-0.5em }
\label{tab:aerialmegadepth_colmap_per_scene}
\end{table*}

Tab.~\ref{tab:xdg_ablation_additional} reports additional controlled ablations of backbone capacity, training augmentations, and bidirectional processing. Each variant changes only the named factor relative to full XDG.

\paragraph{Backbone capacity.}
DA3-Small roughly halves XDG's latency but reduces accuracy, especially on DG. DA3-Large provides a small gain on DG and improves several VisymScenes operating-point metrics, but increases latency by approximately $2.4\times$ over DA3-Base. These results support DA3-Base as the best overall accuracy--efficiency compromise.

\paragraph{Training strategy.}
Removing either random input-order flipping or multi-resolution sampling consistently lowers AP and ROC AUC on both test sets. The degradation from removing multi-resolution training is particularly visible on VisymScenes, indicating that scale and aspect-ratio diversity during training improves generalization.

\paragraph{Order robustness.}
Processing only one image order reduces latency from 34.2 to 19.7~ms on DG and from 33.5 to 19.6~ms on VisymScenes. However, it lowers AP and ROC AUC on both datasets relative to the full bidirectional model. Fig.~\ref{fig:order_robustness} illustrates the practical effect on an example pair. XDG predicts a matching confidence of 0.9727 in both directions, whereas the single-order model's confidence changes from 0.7734 for $A{\rightarrow}B$ to 0.8906 for $B{\rightarrow}A$. At a threshold of 0.8, this difference causes the same candidate edge to be rejected or retained solely because of the arbitrary input order. We therefore retain bidirectional processing to make predictions order-robust, accepting its additional latency.

\vspace{-0.5 em}
\section{Detailed Scene and Sequence Results}

\subsection{AerialMegaDepth: Per-Scene Results}

Tab.~\ref{tab:aerialmegadepth_colmap_per_scene} provides detailed per-scene COLMAP results and visual-disambiguation runtime for AerialMegaDepth.

The gains are concentrated on scenes where repeated structures cause severe ambiguity, most notably Cologne Cathedral and the Louvre Museum. On easier scenes, all three configurations are similar. Across scenes, XDG closely tracks DG++ while requiring substantially less visual-disambiguation time.

\subsection{WRIVA: Per-Sequence Results}

\vspace{-0.5 em}

Tab.~\ref{tab:wriva_gluemap_per_sequence} provides the complete per-sequence GLUEMAP results underlying the aggregate WRIVA numbers in the main paper. We abbreviate each sequence by its unique task--view--scene--run identifier; the number of input images is shown in parentheses.

\begin{table*}[t]
\centering
\scriptsize
\renewcommand{\arraystretch}{0.92}
\setlength{\tabcolsep}{1.5pt}
\begin{tabular}{@{}>{\raggedright\arraybackslash}m{0.16\textwidth}
                *{3}{>{\centering\arraybackslash}m{0.205\textwidth}}
                *{2}{>{\centering\arraybackslash}m{0.065\textwidth}}@{}}
\hline
\noalign{\vskip 2pt}
Sequence (\# images) & No disamb. & DG++ & XDG &
\multicolumn{2}{c}{Time (min)$\downarrow$} \\
\cmidrule(lr){5-6}
& & & & DG++ & XDG \\
\noalign{\vskip 2pt}
\hline
t04-v10-s01-r01 (600) & $2.82/8.81/25.41/52.73$ & $14.69/27.29/52.25/82.88$ & $25.56/45.45/69.93/89.21$ & 62.43 & 19.59 \\
t09-v03-s03-r05 (177) & $1.71/2.91/6.17/14.28$ & $9.68/13.50/19.57/30.16$ & $8.29/12.49/18.87/28.78$ & 16.15 & 5.09 \\
t01-v01-s04-r08 (50) & $60.99/73.07/82.51/91.24$ & $59.05/70.09/81.20/93.22$ & $65.43/78.99/89.36/96.45$ & 1.74 & 0.56 \\
t02-v02-s00-r02 (25) & $41.52/55.49/71.78/90.17$ & $44.89/57.77/68.30/75.35$ & $39.03/57.17/75.67/91.41$ & 0.44 & 0.14 \\
t03-v03-s04-r02 (500) & $3.19/7.86/18.46/40.02$ & $44.60/60.81/77.88/91.72$ & $42.66/59.47/77.33/91.55$ & 33.90 & 16.51 \\
t09-v04-s01-r10 (182) & $2.98/6.63/14.83/36.85$ & $7.85/17.29/38.47/69.98$ & $13.25/28.14/55.56/84.27$ & 17.34 & 5.61 \\
t04-v05-s03-r02 (150) & $8.67/12.92/21.77/53.56$ & $8.44/11.59/18.26/34.03$ & $12.38/20.93/41.31/76.84$ & 12.57 & 3.90 \\
t02-v03-s00-r02 (17) & $14.06/25.79/49.13/77.65$ & $13.30/24.67/48.35/77.39$ & $13.25/24.51/48.26/77.36$ & 0.19 & 0.06 \\
t09-v05-s00-r02 (474) & $0.68/1.22/2.63/8.85$ & $15.16/23.14/38.38/66.47$ & $16.90/24.86/38.45/69.08$ & 45.60 & 14.47 \\
t09-v03-s02-r05 (195) & $0.61/0.82/1.55/6.43$ & $6.80/9.45/13.41/21.12$ & $6.33/9.29/14.96/28.57$ & 18.07 & 6.00 \\
t03-v04-s00-r02 (150) & $12.21/29.00/56.12/82.49$ & $23.99/38.05/61.83/86.05$ & $24.26/38.03/61.64/85.86$ & 6.20 & 4.12 \\
t04-v09-s00-r04 (39) & $3.85/12.06/34.42/63.44$ & $5.27/15.56/34.13/56.20$ & $3.32/8.58/19.84/43.11$ & 0.61 & 0.34 \\
t09-v05-s01-r02 (444) & $0.44/0.75/1.59/5.20$ & $14.89/20.83/31.94/57.97$ & $18.31/25.28/37.35/58.86$ & 42.06 & 13.38 \\
t09-v04-s00-r10 (198) & $1.91/4.58/12.28/34.48$ & $8.93/20.24/42.27/71.77$ & $13.09/27.43/53.47/82.05$ & 19.25 & 6.12 \\
t09-v06-s03-r02 (167) & $20.71/30.56/44.88/66.92$ & $43.02/56.02/68.44/81.51$ & $39.19/53.50/67.28/80.51$ & 14.97 & 4.78 \\
t04-v07-s01-r02 (50) & $2.19/3.03/8.93/48.04$ & $3.12/4.09/6.45/15.47$ & $2.90/2.95/3.16/6.19$ & 1.22 & 0.56 \\
t09-v05-s02-r02 (414) & $0.59/0.96/1.81/5.85$ & $18.76/27.38/40.78/68.19$ & $9.96/13.44/18.62/31.14$ & 39.38 & 12.66 \\
t03-v05-s02-r04 (200) & $17.58/20.40/23.27/28.45$ & $26.34/31.01/35.71/42.02$ & $26.40/31.26/35.96/41.63$ & 16.93 & 5.31 \\
t09-v04-s02-r10 (174) & $2.67/5.29/11.89/33.29$ & $9.12/20.77/43.05/73.09$ & $12.28/24.87/49.40/76.83$ & 16.48 & 5.22 \\
t02-v02-s01-r02 (50) & $46.69/60.07/73.70/87.53$ & $52.69/65.21/75.09/83.10$ & $49.96/61.05/71.17/81.09$ & 1.74 & 0.55 \\
t01-v01-s07-r08 (10) & $55.77/73.46/86.73/95.58$ & $44.38/62.67/81.30/93.77$ & $45.18/60.44/79.51/93.17$ & 0.07 & 0.02 \\
t09-v05-s03-r02 (396) & $0.92/1.50/2.52/7.46$ & $19.12/26.66/39.17/63.03$ & $11.09/14.39/18.89/33.16$ & 37.15 & 11.73 \\
t09-v03-s01-r05 (225) & $0.60/0.81/1.56/6.82$ & $4.75/7.10/10.68/16.26$ & $6.16/9.81/15.58/24.49$ & 21.56 & 6.88 \\
t09-v06-s02-r02 (170) & $4.77/11.19/26.42/55.94$ & $38.80/52.63/66.29/80.84$ & $30.86/46.92/63.08/79.32$ & 15.20 & 4.77 \\
t02-v05-s01-r02 (50) & $28.90/46.24/62.40/80.10$ & $29.92/50.14/68.57/81.86$ & $16.42/36.03/58.35/80.56$ & 1.77 & 0.55 \\
t01-v01-s08-r08 (5) & $41.57/62.48/81.24/93.75$ & $34.87/58.60/79.30/93.10$ & $27.68/37.16/50.73/59.58$ & 0.01 & 0.01 \\
t04-v01-s00-r06 (300) & $2.17/4.90/13.02/35.11$ & $20.54/30.91/41.81/53.79$ & $21.47/33.89/48.49/66.86$ & 28.65 & 8.99 \\
t04-v10-s00-r01 (300) & $2.24/4.66/10.51/27.18$ & $14.16/23.18/36.13/50.89$ & $12.48/20.65/35.34/64.80$ & 19.55 & 9.22 \\
t09-v06-s01-r02 (175) & $20.98/26.66/37.78/67.47$ & $37.68/51.20/64.53/78.81$ & $34.22/48.79/63.45/78.07$ & 16.15 & 5.03 \\
t02-v05-s02-r02 (75) & $3.52/7.84/19.20/46.96$ & $13.55/21.55/30.57/50.45$ & $14.14/21.99/31.89/53.01$ & 3.95 & 1.26 \\
t04-v01-s01-r06 (300) & $1.21/2.44/6.22/21.78$ & $5.80/9.04/13.53/21.13$ & $8.67/13.19/21.99/44.75$ & 27.03 & 8.51 \\
t09-v03-s00-r05 (255) & $0.58/1.06/3.14/11.56$ & $3.83/5.76/8.88/16.02$ & $4.64/7.94/14.55/28.90$ & 25.37 & 7.83 \\
t09-v06-s00-r02 (180) & $34.86/49.12/65.97/84.67$ & $35.17/48.28/61.58/77.66$ & $31.11/45.16/59.68/76.37$ & 19.19 & 5.10 \\
t02-v05-s04-r02 (107) & $3.23/7.92/21.85/53.72$ & $16.94/23.69/36.60/60.33$ & $15.18/21.38/35.42/63.02$ & 8.28 & 2.55 \\
\hline
\end{tabular}
\caption{\textbf{Per-sequence GLUEMAP reconstruction on WRIVA.} Each AUC cell reports pose AUC (\%) as the tuple AUC@$3^{\circ}$ / @$5^{\circ}$ / @$10^{\circ}$ / @$30^{\circ}$. Time is the visual-disambiguation two-view inference time for each sequence. Results cover all 34 evaluation sequences.}
\label{tab:wriva_gluemap_per_sequence}
\end{table*}

The sequence-level results reveal substantial variation across capture conditions. Visual disambiguation provides large gains on many difficult PTZ and varying-altitude sequences, while a few small or already well-reconstructed sequences favor no filtering. XDG and DG++ exhibit similar overall behavior, with complementary strengths across individual sequences.

\section{More Qualitative Results}

Figs.~\ref{fig:aerial_qualitative_adler_belfry}--\ref{fig:aerial_qualitative_louvre_ponte} provide additional top-down visualizations of AerialMegaDepth COLMAP reconstructions. For each scene, we show the ground-truth camera layout and reconstructions without visual disambiguation, with DG++, and with XDG. Red frusta denote reconstructed cameras. When a method produces multiple substantial reconstruction components, we stack the aligned components vertically within the same column.

Figs.~\ref{fig:wriva_qualitative_trailer}--\ref{fig:wriva_qualitative_singlezone_altitude} show six representative WRIVA GLUEMAP reconstructions that are distinct from the two examples in the main paper. We select sequences for which both DG++ and XDG consistently outperform reconstruction without visual disambiguation. Across these examples, visual disambiguation removes geometrically inconsistent matches and yields camera layouts that better agree with the ground truth.

\begingroup
\newcommand{\aerialvis}[1]{%
    \includegraphics[width=\linewidth]{aerialmegadepth_colmap/#1}%
}
\newcommand{\aerialvisstack}[2]{%
    \begin{minipage}[c]{\linewidth}
        \centering
        \aerialvis{#1}\par\vspace{1pt}
        \aerialvis{#2}
    \end{minipage}%
}

\begin{figure*}[p]
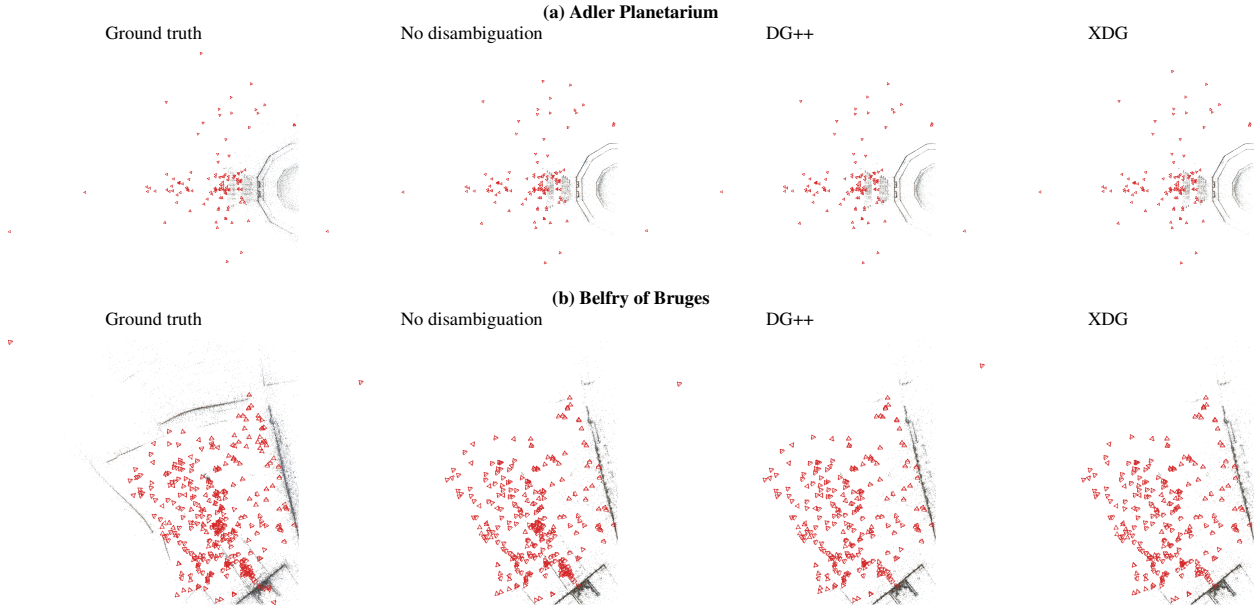

\centering
\scriptsize
\setlength{\tabcolsep}{1.5pt}
\begin{tabular}{*{4}{>{\centering\arraybackslash}m{0.235\textwidth}}}
\multicolumn{4}{c}{\textbf{(a) Adler Planetarium}} \\
Ground truth & No disambiguation & DG++ & XDG \\
\aerialvis{adler_planetarium/ground_truth.png} &
\aerialvis{adler_planetarium/no_disambiguation/0.png} &
\aerialvis{adler_planetarium/dgpp/0.png} &
\aerialvis{adler_planetarium/xdg/0.png} \\
\noalign{\vskip 5pt}
\multicolumn{4}{c}{\textbf{(b) Belfry of Bruges}} \\
Ground truth & No disambiguation & DG++ & XDG \\
\aerialvis{belfry_of_bruges/ground_truth.png} &
\aerialvis{belfry_of_bruges/no_disambiguation/0.png} &
\aerialvis{belfry_of_bruges/dgpp/0.png} &
\aerialvis{belfry_of_bruges/xdg/1.png} \\
\end{tabular}
\caption{\textbf{Additional AerialMegaDepth COLMAP reconstructions (1/3).} Ground-truth and reconstructed camera layouts for Adler Planetarium and Belfry of Bruges.}
\label{fig:aerial_qualitative_adler_belfry}
\end{figure*}

\begin{figure*}[p]
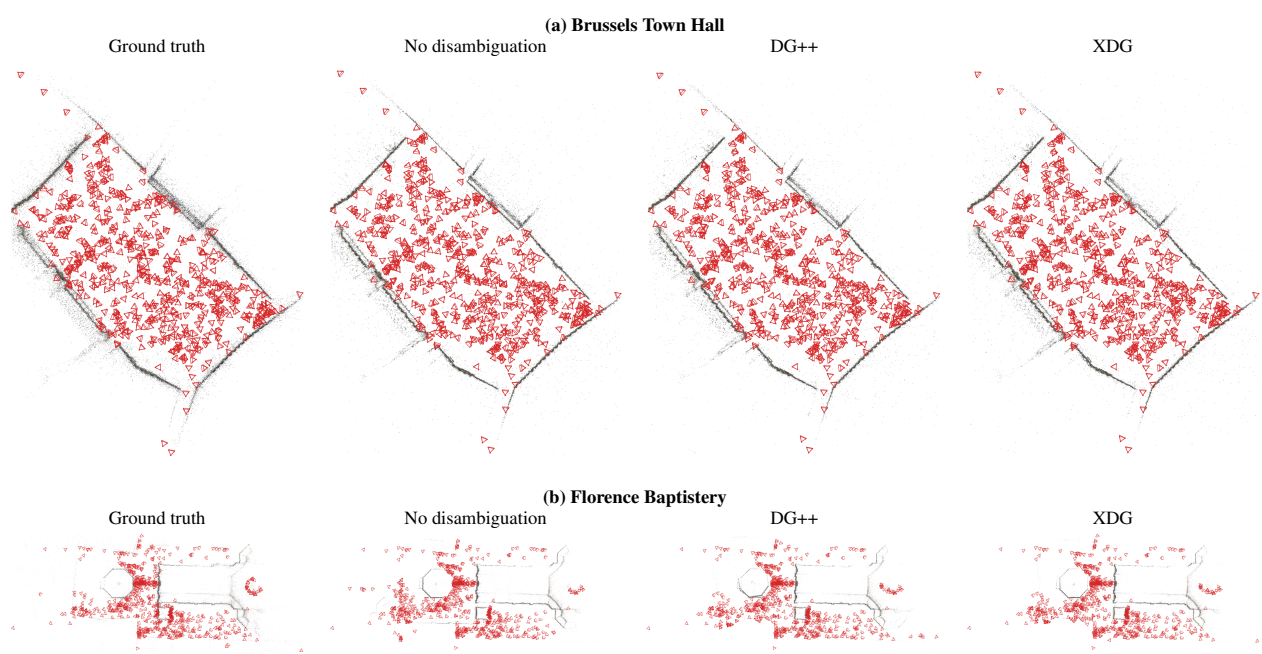

\centering
\scriptsize
\setlength{\tabcolsep}{1.5pt}
\begin{tabular}{*{4}{>{\centering\arraybackslash}m{0.235\textwidth}}}
\multicolumn{4}{c}{\textbf{(a) Brussels Town Hall}} \\
Ground truth & No disambiguation & DG++ & XDG \\
\aerialvis{brussels_town_hall/ground_truth.png} &
\aerialvis{brussels_town_hall/no_disambiguation/0.png} &
\aerialvis{brussels_town_hall/dgpp/0.png} &
\aerialvis{brussels_town_hall/xdg/0.png} \\
\noalign{\vskip 5pt}
\multicolumn{4}{c}{\textbf{(b) Florence Baptistery}} \\
Ground truth & No disambiguation & DG++ & XDG \\
\aerialvis{florence_baptistery/ground_truth.png} &
\aerialvis{florence_baptistery/no_disambiguation/0.png} &
\aerialvis{florence_baptistery/dgpp/0.png} &
\aerialvis{florence_baptistery/xdg/0.png} \\
\end{tabular}
\caption{\textbf{Additional AerialMegaDepth COLMAP reconstructions (2/3).} Ground-truth and reconstructed camera layouts for Brussels Town Hall and Florence Baptistery.}
\label{fig:aerial_qualitative_brussels_florence}
\end{figure*}

\begin{figure*}[p]
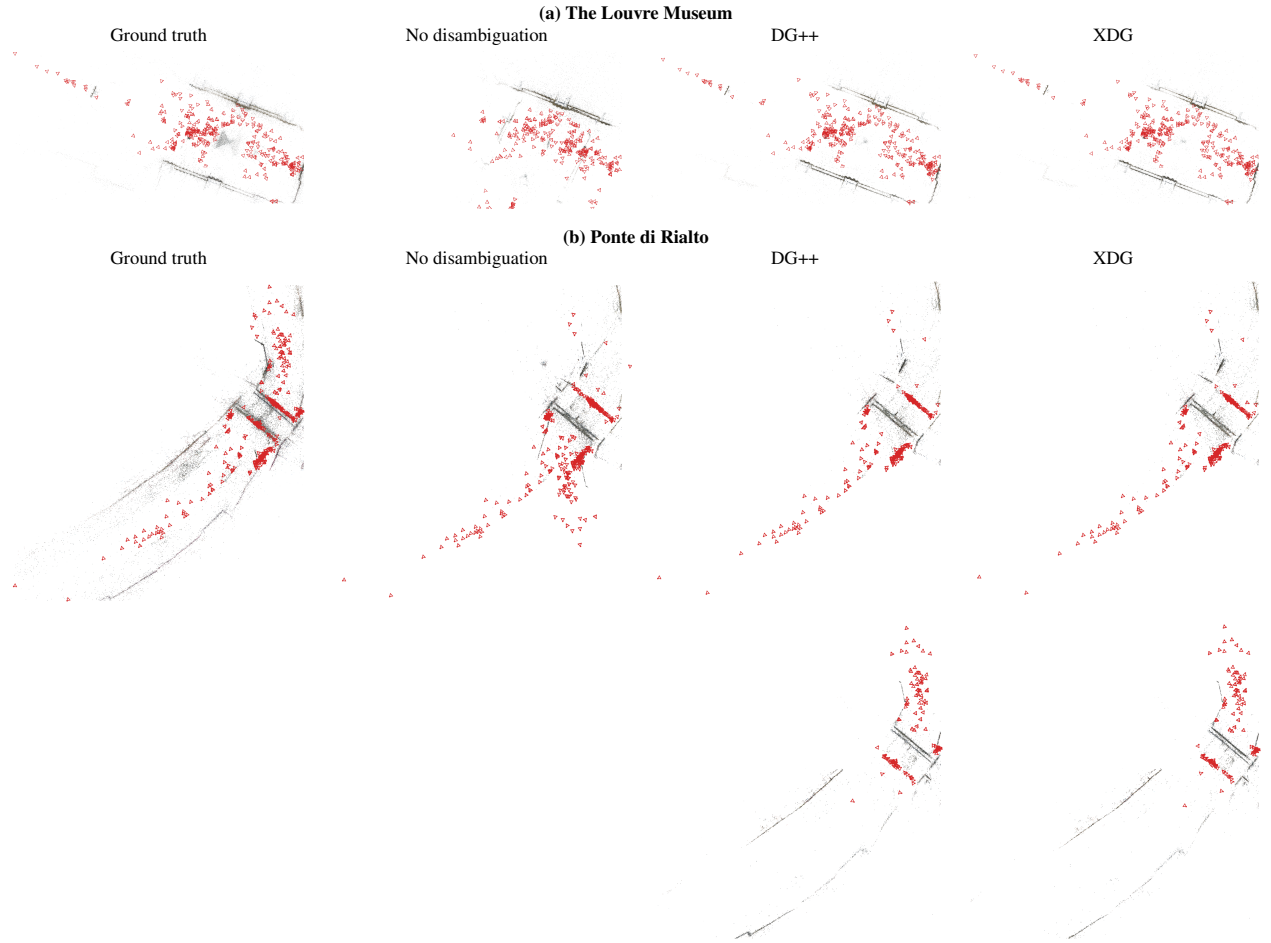

\centering
\scriptsize
\setlength{\tabcolsep}{1.5pt}
\begin{tabular}{*{4}{>{\centering\arraybackslash}m{0.235\textwidth}}}
\multicolumn{4}{c}{\textbf{(a) The Louvre Museum}} \\
Ground truth & No disambiguation & DG++ & XDG \\
\aerialvis{the_louvre_museum/ground_truth.png} &
\aerialvis{the_louvre_museum/no_disambiguation/0.png} &
\aerialvis{the_louvre_museum/dgpp/0.png} &
\aerialvis{the_louvre_museum/xdg/1.png} \\
\noalign{\vskip 5pt}
\multicolumn{4}{c}{\textbf{(b) Ponte di Rialto}} \\
Ground truth & No disambiguation & DG++ & XDG \\
\aerialvis{ponte_di_rialto/ground_truth.png} &
\aerialvis{ponte_di_rialto/no_disambiguation/0.png} &
\aerialvisstack{ponte_di_rialto/dgpp/0.png}
                {ponte_di_rialto/dgpp/1.png} &
\aerialvisstack{ponte_di_rialto/xdg/0.png}
                {ponte_di_rialto/xdg/1.png} \\
\end{tabular}
\caption{\textbf{Additional AerialMegaDepth COLMAP reconstructions (3/3).} Ground-truth and reconstructed camera layouts for the Louvre Museum and Ponte di Rialto. For Ponte di Rialto, DG++ and XDG each produce two reconstruction components, which are stacked vertically.}
\label{fig:aerial_qualitative_louvre_ponte}
\end{figure*}

\newcommand{\wrivavis}[1]{%
    \includegraphics[width=\linewidth]{wriva_gluemap/#1}%
}

\begin{figure*}[p]
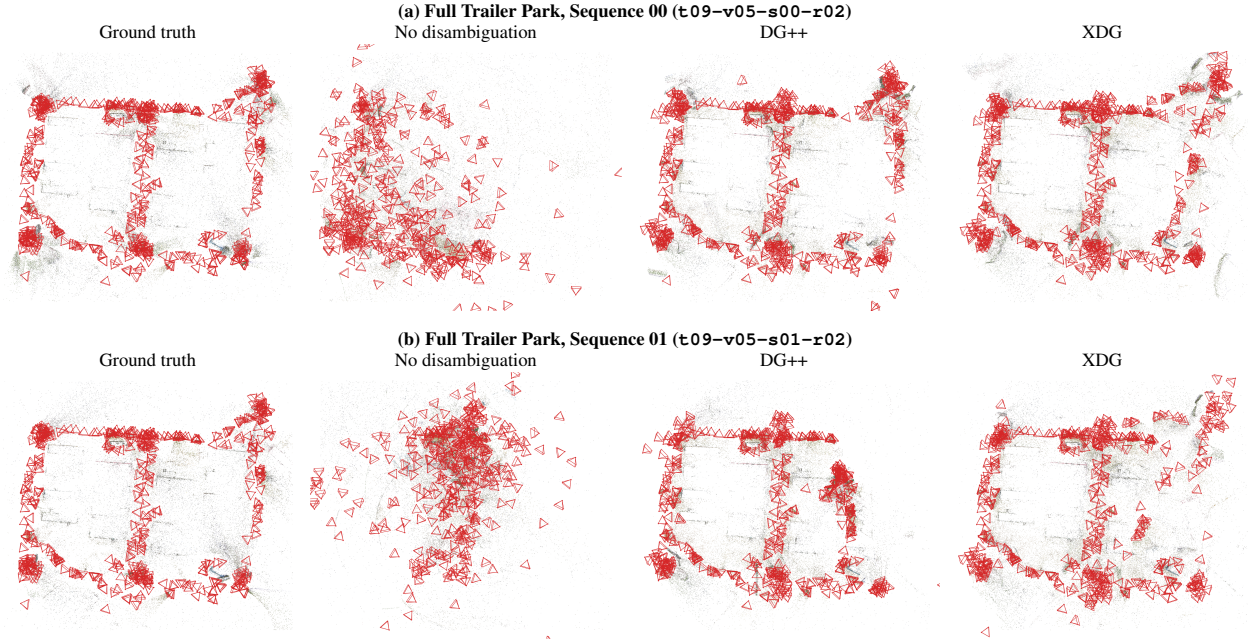

\centering
\scriptsize
\setlength{\tabcolsep}{1.5pt}
\begin{tabular}{*{4}{>{\centering\arraybackslash}m{0.235\textwidth}}}
\multicolumn{4}{c}{\textbf{(a) Full Trailer Park, Sequence 00 (\texttt{t09-v05-s00-r02})}} \\
Ground truth & No disambiguation & DG++ & XDG \\
\wrivavis{full_trailer_park/ground_truth.png} &
\wrivavis{full_trailer_park/no_disambiguation.png} &
\wrivavis{full_trailer_park/dgpp.png} &
\wrivavis{full_trailer_park/xdg.png} \\
\noalign{\vskip 5pt}
\multicolumn{4}{c}{\textbf{(b) Full Trailer Park, Sequence 01 (\texttt{t09-v05-s01-r02})}} \\
Ground truth & No disambiguation & DG++ & XDG \\
\wrivavis{full_trailer_park_s01/ground_truth.png} &
\wrivavis{full_trailer_park_s01/no_disambiguation.png} &
\wrivavis{full_trailer_park_s01/dgpp.png} &
\wrivavis{full_trailer_park_s01/xdg.png} \\
\end{tabular}
\caption{\textbf{Representative WRIVA GLUEMAP reconstructions (1/3).} On Sequence 00, AUC@$10^{\circ}$ increases from 2.63\% without disambiguation to 38.38\% with DG++ and 38.45\% with XDG. On Sequence 01, it increases from 1.59\% to 31.94\% and 37.35\%, respectively.}
\label{fig:wriva_qualitative_trailer}
\end{figure*}

\begin{figure*}[p]
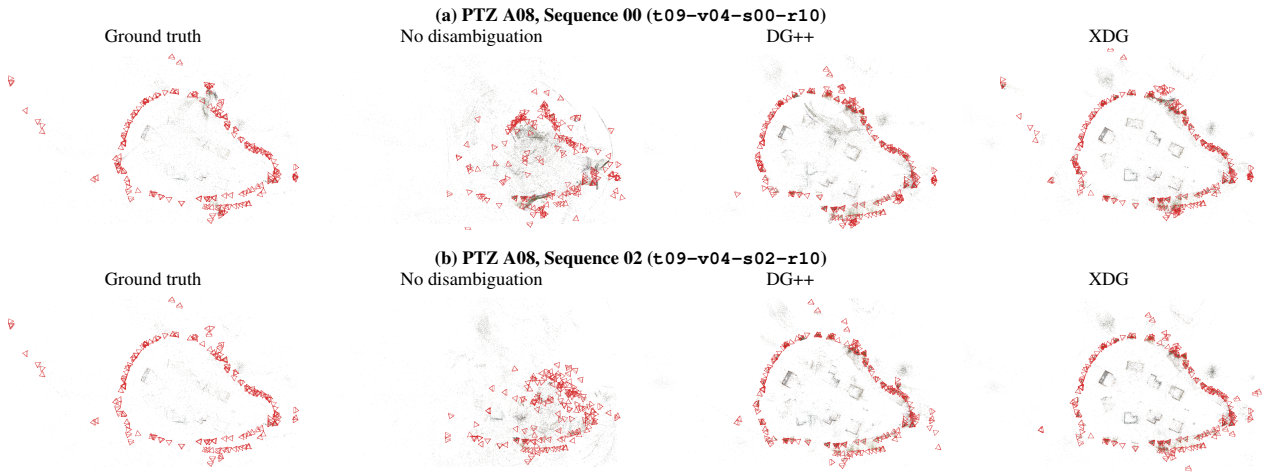

\centering
\scriptsize
\setlength{\tabcolsep}{1.5pt}
\begin{tabular}{*{4}{>{\centering\arraybackslash}m{0.235\textwidth}}}
\multicolumn{4}{c}{\textbf{(a) PTZ A08, Sequence 00 (\texttt{t09-v04-s00-r10})}} \\
Ground truth & No disambiguation & DG++ & XDG \\
\wrivavis{ptz_a08_s00/ground_truth.png} &
\wrivavis{ptz_a08_s00/no_disambiguation.png} &
\wrivavis{ptz_a08_s00/dgpp.png} &
\wrivavis{ptz_a08_s00/xdg.png} \\
\noalign{\vskip 5pt}
\multicolumn{4}{c}{\textbf{(b) PTZ A08, Sequence 02 (\texttt{t09-v04-s02-r10})}} \\
Ground truth & No disambiguation & DG++ & XDG \\
\wrivavis{ptz_a08_s02/ground_truth.png} &
\wrivavis{ptz_a08_s02/no_disambiguation.png} &
\wrivavis{ptz_a08_s02/dgpp.png} &
\wrivavis{ptz_a08_s02/xdg.png} \\
\end{tabular}
\caption{\textbf{Representative WRIVA GLUEMAP reconstructions (2/3).} On Sequence 00, AUC@$10^{\circ}$ increases from 12.28\% without disambiguation to 42.27\% with DG++ and 53.47\% with XDG. On Sequence 02, it increases from 11.89\% to 43.05\% and 49.40\%, respectively.}
\label{fig:wriva_qualitative_a08}
\end{figure*}

\begin{figure*}[p]
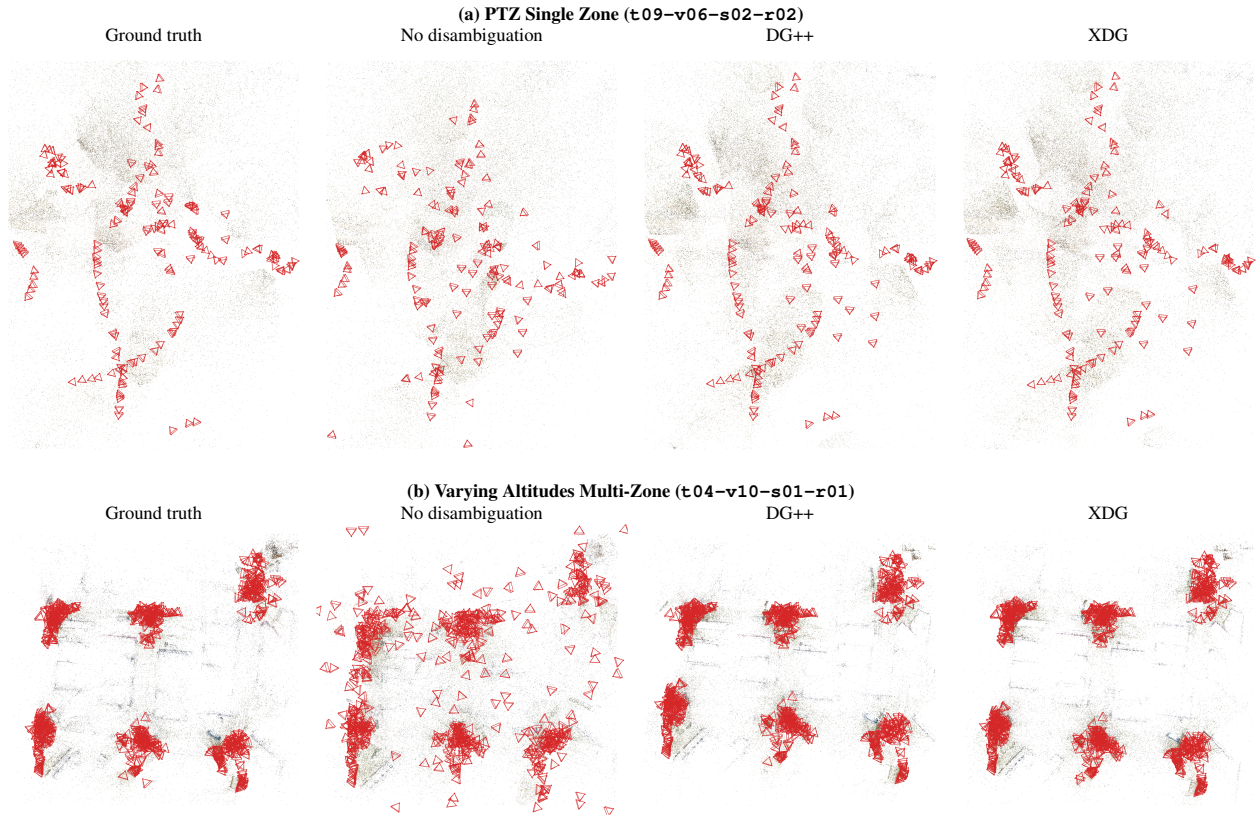

\centering
\scriptsize
\setlength{\tabcolsep}{1.5pt}
\begin{tabular}{*{4}{>{\centering\arraybackslash}m{0.235\textwidth}}}
\multicolumn{4}{c}{\textbf{(a) PTZ Single Zone (\texttt{t09-v06-s02-r02})}} \\
Ground truth & No disambiguation & DG++ & XDG \\
\wrivavis{ptz_single_zone/ground_truth.png} &
\wrivavis{ptz_single_zone/no_disambiguation.png} &
\wrivavis{ptz_single_zone/dgpp.png} &
\wrivavis{ptz_single_zone/xdg.png} \\
\noalign{\vskip 5pt}
\multicolumn{4}{c}{\textbf{(b) Varying Altitudes Multi-Zone (\texttt{t04-v10-s01-r01})}} \\
Ground truth & No disambiguation & DG++ & XDG \\
\wrivavis{varying_altitudes_multi_zone/ground_truth.png} &
\wrivavis{varying_altitudes_multi_zone/no_disambiguation.png} &
\wrivavis{varying_altitudes_multi_zone/dgpp.png} &
\wrivavis{varying_altitudes_multi_zone/xdg.png} \\
\end{tabular}
\caption{\textbf{Representative WRIVA GLUEMAP reconstructions (3/3).} On PTZ Single Zone, AUC@$10^{\circ}$ increases from 26.42\% without disambiguation to 66.29\% with DG++ and 63.08\% with XDG. On Varying Altitudes Multi-Zone, it increases from 25.41\% to 52.25\% and 69.93\%, respectively.}
\label{fig:wriva_qualitative_singlezone_altitude}
\end{figure*}

\endgroup

\end{document}